%% file: main.tex
\pdfoutput=1
\documentclass[11pt]{article}

\usepackage[preprint]{acl}

\usepackage{times}
\usepackage{latexsym}
\usepackage[T1]{fontenc}
\usepackage[utf8]{inputenc}
\usepackage{microtype}
\usepackage{inconsolata}
\usepackage{graphicx}
\usepackage{booktabs}
\usepackage{multirow}
\usepackage{xcolor}
\usepackage{enumitem}

\makeatletter
\newcommand{\releaselink}{\ifacl@anonymize\else\footnote{\url{https://huggingface.co/collections/hafsteinn/icelandic-modernbert}}\fi}
\makeatother

\title{Beyond IceBERT: Open-Data ModernBERT Encoders for Icelandic from 12M to 1B Parameters}

\author{Hafsteinn Einarsson \\ University of Iceland}

\begin{document}
\maketitle

\begin{abstract}
IceBERT, the reference encoder for Icelandic, found that a cleaned web crawl is enough to train a strong model and that starting from a multilingual model can help on some tasks.
We revisit both findings with the ModernBERT architecture.
We train Icelandic encoders from 12M to 1B parameters on 9B tokens of open data, and rerun IceBERT's benchmarks and five newer tasks for 30 encoders under one protocol with confidence intervals.
Our 1B model trained from scratch, to our knowledge the first monolingual Icelandic encoder of that size, is significantly better than IceBERT on all eleven tasks; warm-started from the English Ettin encoder of the same shape, it has the best score of all encoders on every task we rank.
Because our model shapes match the English Ettin encoders and the multilingual mmBERT, we can also test the warm start under controlled conditions, at six model sizes and four token budgets.
With our vocabulary transplanted, a warm start was never significantly worse than scratch training on the same Icelandic text, and from base size up it was not significantly worse on four tasks than scratch training on 4 times the text (8 times at base and 1B).
As models grow, most of the transfer comes from the source model's transformer layers, so even an English source helps.
Keeping the source vocabulary instead cost 5 F1 on question answering and nearly three times the compute.
\end{abstract}

\section{Introduction}
IceBERT \citep{snaebjarnarson-etal-2022-warm} has been the reference encoder for Icelandic since 2022.
Its paper made two claims: a cleaned web crawl is enough to train a strong Icelandic model, and starting from a multilingual model can give state-of-the-art results on some tasks.
The warm-started model, XLM-R base \citep{conneau-etal-2020-unsupervised} continued on the Icelandic crawl, was their best model on NER and used about 8\% of IceBERT's GPU hours.

Much has changed since.
ModernBERT \citep{warner-etal-2025-smarter} renewed the encoder architecture and training recipe.
Ettin \citep{weller-etal-2026-seq} released a ladder of open encoders of the same design from 17M to 1B parameters, trained on English and code, and mmBERT \citep{marone-etal-2026-mmbert} trained the design on more than 1,800 languages, Icelandic among them.
Open Icelandic text has grown as well: the sources we use give 9B tokens after deduplication, against 2.7B for IceBERT.

We ask two questions.
First, how far does a monolingual ModernBERT trained only on open data get on IceBERT's benchmarks?
Second, does IceBERT's warm-start finding hold when the comparison is controlled, which source model is best, and how much Icelandic text does a warm start save?
IceBERT's own comparison could not settle the second question, since its warm-started model differed from its scratch models in vocabulary, training data, size and training length.
Because our models follow the Ettin shapes, each has an English source model of exactly the same shape, and our base model also has a multilingual one, so a warm start changes nothing but the starting weights.

Our contributions are:
\begin{itemize}[nosep,leftmargin=*]
\item Icelandic ModernBERT encoders from 12M to 1B parameters, trained only on open data (five sizes up to 378M at both 80 and 160 tokens per parameter, and 1B at 80), which we release together with the warm-started models\releaselink{} (Sections~\ref{sec:data} and~\ref{sec:models});
\item IceBERT's benchmarks and five newer tasks rerun for 30 encoders under one protocol, with learning rates tuned until the choice is interior and bootstrap confidence intervals (Section~\ref{sec:eval}; all 30 in Appendix~\ref{app:ci});
\item a controlled test of the warm start: scratch training against English and multilingual warm starts of identical shape, at six sizes and four token budgets, with an arm that separates the transformer layers from the embeddings (Section~\ref{sec:warm}).
\end{itemize}
Our 1B model trained from scratch is significantly better than IceBERT on all eleven tasks, and warm-started from Ettin it has the best mean of all encoders on every ranked task.
A warm start with our vocabulary was never significantly worse than scratch training on the same Icelandic text, and from base size up it was not significantly worse on four tasks than scratch training on four times the text, and at base and 1B on eight times.

\section{Related work}

\paragraph{Icelandic and Nordic encoders.}
IceBERT trained RoBERTa-base and -large models on 16GB of Icelandic text \citep{snaebjarnarson-etal-2022-warm}.
\citet{gudnason-loftsson-2022-pre} trained ELECTRA and ConvBERT models on the Icelandic Gigaword Corpus and found that they beat a multilingual model of similar size; we evaluate both at base and small size.
ScandiBERT was trained on all the Scandinavian languages, Icelandic and Faroese included \citep{snaebjarnarson-etal-2023-transfer}.
For neighbouring languages, NorBench introduced a ladder of Norwegian encoders \citep{samuel-etal-2023-norbench}, and Finnish ModernBERTs were strongest on long inputs but behind XLM-R on most short-context tasks \citep{reunamo-etal-2025-pretraining}.
EuroBERT \citep{boizard-etal-2025-eurobert} covers 15 languages, none of them Nordic.

\paragraph{Icelandic evaluation.}
ScandEval \citep{nielsen-2023-scandeval} benchmarks four tasks in the Scandinavian languages, Icelandic included, introduced the acceptability and QA datasets, and found limited transfer between the mainland languages and Icelandic and Faroese; on such tasks, encoders can beat decoders that are orders of magnitude larger \citep{saattrup-nielsen-etal-2025-encoder}.
Icelandic QA data was written by native speakers in NQiI \citep{snaebjarnarson-einarsson-2022-natural} and crowd-sourced through a mobile game for RUQuAD \citep{skarphedinsson-etal-2023-gameqa}.
An audit of Icelandic benchmarks found machine-translated test sets to be severely flawed, while Icelandic WinoGrande had no severely flawed items \citep{ingimundarson-etal-2026-benchmarks}; every task we use was written in Icelandic or translated by hand.

\paragraph{Warm starts and vocabulary transfer.}
Continued pretraining adapts a model to a domain \citep{gururangan-etal-2020-dont} or to a language \citep{chau-etal-2020-parsing,ralethe-2020-adaptation}.
When the target language gets its own tokenizer, the embeddings must be carried over: Fast Vocabulary Transfer averages the source embeddings of each new token's pieces \citep{gee-etal-2022-fast}, while WECHSEL \citep{minixhofer-etal-2022-wechsel} and FOCUS \citep{dobler-de-melo-2023-focus} use static word embeddings or overlapping tokens.
Relearning only the embeddings of English GPT-2 already adapts it to Italian and Dutch, which suggests the transformer layers are partly language-independent \citep{de-vries-nissim-2021-good}.
How much target-language data a warm start saves has been measured for decoders.
\citet{hernandez-etal-2021-scaling} define the effective data transferred as the extra data a scratch model would need for the same loss.
WECHSEL matched scratch models with up to 64 times less training, CLP-Transfer reached their perplexity with 20--50\% of the tokens \citep{ostendorff-rehm-2023-efficient}, continued pretraining from English to Chinese saved 25--50\% \citep{zheng-etal-2024-breaking}, and ATLAS gives the budgets at which training from scratch overtakes a multilingual checkpoint \citep{longpre-etal-2026-atlas}.
For encoders, XLM-R continued on Croatian and related languages reached the level of a model trained from scratch \citep{ljubesic-etal-2024-language}.
We know of no encoder study that measures the saving across several model sizes and target-language budgets, or that compares an English and a multilingual source of the same shape.

\section{Data}
\label{sec:data}
We use only text that is openly available.
We use nine sources (Table~\ref{tab:data} in Appendix~\ref{app:data}): the open part of the Icelandic Gigaword Corpus \citep{barkarson-etal-2022-evolving}, Icelandic Wikipedia \citep{wikimedia-2023-wikipedia}, the parts of Icelandic Dynaword that are not taken from the Gigaword Corpus \citep{enevoldsen-etal-2026-dynaword,dfm-2026-icelandic-dynaword}, MaCoCu \citep{non-etal-2022-macocu}, FineWeb-2 \citep{penedo-etal-2025-fineweb2}, FinePDFs \citep{kydlicek-etal-2025-finepdfs}, HPLT~3.0 \citep{oepen-etal-2026-hplt}, MADLAD-400 \citep{kudugunta-etal-2023-madlad} and IC3-v2 \citep{mideind-2024-ic3v2}.
The restricted parts of the Gigaword Corpus (news and books under separate agreements) are not used.

Every source goes through the same four steps; Appendix~\ref{app:data} gives the thresholds.
A filter removes short documents, documents whose function words point to Faroese, a Scandinavian language or English, and documents with too much repetition or boilerplate.
Exact and MinHash deduplication then run across all sources at once, keeping the copy from the more curated source.
Decontamination removes every line that shares a word 13-gram, or for sentences of 6 to 12 words the whole sentence, with any split of the evaluation sets we used at that point: NER, UD PoS tagging and acceptability, sentence-level error detection and QA (the 13-gram rule follows \citealp{brown-etal-2020-language}); a document that would lose more than a fifth of its words is dropped.
Appendix~\ref{app:harness} gives the overlaps for the tasks added later.
Finally 0.1\% of the documents of each source are held out for validation.

The result is 9.02B tokens.
Most of the web data overlaps: 84\% of FineWeb-2's documents and about two thirds of HPLT's, MADLAD's and IC3's have a near-duplicate elsewhere, while the Gigaword Corpus and FinePDFs contribute most of the text that no other source has.
Decontamination dropped 64,401 documents and shortened 111,092; the NER (MIM-GOLD-NER) and UD PoS sentences were the most frequent matches.
For the models trained on twice the tokens we add the part of CulturaX \citep{nguyen-etal-2024-culturax} that survives the same pipeline, deduplicated last so that the other sources keep every document, 0.34B tokens (+3.7\%); 77\% of CulturaX's Icelandic documents already occur in the other sources.

\section{Models}
\label{sec:models}

The architecture is ModernBERT \citep{warner-etal-2025-smarter}: pre-normalisation without biases, GeGLU feed-forward layers, rotary position embeddings, and global attention in every third layer with a 128-token sliding window in the others.
Model shapes follow the Ettin ladder \citep{weller-etal-2026-seq}, whose 150m and 400m sizes are ModernBERT-base and -large, so that each of our models has an English Ettin encoder of exactly the same shape (Section~\ref{sec:warm}).
Table~\ref{tab:sizes} gives the six sizes; the largest, 1B, was added last and not trained at 160 tokens per parameter.
The tokenizer is a byte-level BPE with 32,768 tokens trained on our corpus; it splits held-out text into 1.54 tokens per word, against 1.51 for IceBERT's 50k vocabulary.
A 50k vocabulary trained on our corpus would have been 3.9\% more compact than ours, too little to justify an embedding matrix that would be three quarters of the smallest model.

\begin{table}[t]
\centering
\small
\setlength{\tabcolsep}{5pt}
\begin{tabular}{lrrrrr}
\toprule
Size & Layers & Hidden & Params & Tokens & Node-h \\
\midrule
xs & 7 & 256 & 12.4M & 0.99B & 0.1 \\
s & 10 & 384 & 25.3M & 2.02B & 0.2 \\
m & 19 & 512 & 59.4M & 4.75B & 1.1 \\
base & 22 & 768 & 136.1M & 10.89B & 4.5 \\
large & 28 & 1024 & 377.8M & 30.23B & 25 \\
1B & 28 & 1792 & 999.8M & 79.98B & 135 \\
\bottomrule
\end{tabular}
\caption{Model sizes, with training tokens at 80 tokens per parameter (embeddings included) and node-hours on MareNostrum~5 (four H100 GPUs per node) estimated from measured throughput (1B: 135 node-hours measured, on 16 nodes). The models trained on 160 tokens per parameter see twice the tokens; the 1B model was not trained at 160. Parameter counts are below the names of the Ettin models of the same shape (17M to 1B) because our vocabulary has 32,768 tokens against Ettin's 50,368. Appendix~\ref{app:hyper} gives the hyperparameters.}
\label{tab:sizes}
\end{table}

Each model is trained for 80 tokens per parameter, counting all parameters as \citet{hoffmann-etal-2022-training} do, and a second series for 160.
Training uses masked language modelling at a 30\% masking rate \citep{wettig-etal-2023-mask}, lowered to 15\% for the final decay as in Ettin.
The learning rate follows a warmup--stable--decay schedule \citep{hu-etal-2024-minicpm,hagele-etal-2024-scaling}: 3\% warmup, a constant peak, and a $1-\sqrt{x}$ decay over the last 15\% of the tokens.
The first 90\% of the tokens use 1,024-token sequences and the last 10\% 8,192-token sequences, drawn only from sources whose documents keep their paragraph order (Table~\ref{tab:data}); the rotary settings stay the same throughout.
Documents are packed without padding and kept apart by a document attention mask.
Our implementation reproduces the transformers ModernBERT logits exactly, so every model is released as a standard checkpoint.
With 9.02B unique tokens (9.36B with CulturaX), the large model sees its data 3.4 times at 80 tokens per parameter and 6.5 times at 160, and the 1B model 8.9 times at 80.

\section{A warm start, revisited}
\label{sec:warm}
IceBERT's warm start continued XLM-R, a multilingual model that had seen Icelandic, and kept its vocabulary \citep{snaebjarnarson-etal-2022-warm}.
We ask the same question under controlled conditions.
Because our sizes follow the Ettin ladder, each of them has an English Ettin encoder \citep{weller-etal-2026-seq} with exactly the same layers, width and heads, and mmBERT-base \citep{marone-etal-2026-mmbert}, which saw Icelandic in its later training phases, has exactly our base shape.
A warm start therefore changes only the starting weights; the architecture, data, tokenizer, schedule and evaluation stay those of the scratch models.

We compare five arms.
\textbf{Scratch} trains from random initialisation.
\textbf{Ettin} starts from the Ettin encoder of the same size, with its embeddings moved onto our vocabulary by Fast Vocabulary Transfer \citep{gee-etal-2022-fast}: each of our tokens is re-segmented by Ettin's tokenizer and receives the mean of the embeddings (and output biases) of the pieces.
\textbf{Ettin-body} keeps Ettin's transformer layers but gives our vocabulary freshly initialised embeddings, which separates what the body transfers from what the transplanted embeddings add.
At base size, \textbf{mmBERT-transplant} starts from mmBERT-base with our vocabulary transplanted onto it in the same way, and \textbf{mmBERT-keep} continues mmBERT-base with its own 256k Gemma tokenizer, as IceBERT continued XLM-R.
We use FVT because it needs nothing but the two tokenizers, so the same procedure serves an English and a multilingual source; methods that align embeddings through static word vectors, such as WECHSEL and FOCUS, may carry over more, so the share of the transfer that we attribute to the embeddings is a lower estimate.
mmBERT-keep is matched to the other arms in text rather than in tokens, since the Gemma tokenizer needs 1.60 times as many tokens for our corpus: at the 80 point it trains on 17.46B Gemma tokens, 57 per parameter of its 307M, in 1.6 times as many steps of the same batch size.
Its 128-token attention window and its fine-tuning windows therefore cover about 37\% less text than those of the other arms.
Warm-started arms use the source models' rotary settings and the scratch learning rate; in a pilot at the smallest size, half and a quarter of that rate gave a higher loss for both Ettin and Ettin-body, and at base size half the rate is worse for Ettin and makes no difference for mmBERT-keep (Appendix~\ref{app:checks}).

To measure how many Icelandic tokens a warm start saves, every arm is trained at 10, 20, 40 and 80 tokens per parameter.
A warmup--stable--decay schedule allows this from one run \citep{hagele-etal-2024-scaling}: the 80-token run keeps its state at the start of each shorter budget's decay phase (8.5, 17 and 34 tokens per parameter), and a short decay branch from that state completes the shorter budget.
Each branch is a complete schedule for its own budget: it decays over the last 15\% of that budget, with the lower masking rate, and uses 8,192-token sequences in its last 10\%.
All branches share the main run's warmup, which is 24\% of the 10-token budget, for every arm alike.
At xs, s and m we train Scratch and Ettin with two pretraining seeds each and Ettin-body with one; at base each arm has one seed.
At large and 1B we train Scratch and Ettin with one seed each.
Both scratch curves were trained after the rest of the study: at 1B the branches start from the kept states of the 80-token run, while the ladder's large model kept no such states, so the large scratch curve is a second run of its recipe and seed (see Limitations).

\section{Evaluation}
\label{sec:eval}

We evaluate on the six benchmarks of the IceBERT paper and add five that later became common for Icelandic.
\emph{PoS tagging} uses MIM-GOLD 21.05 \citep{loftsson-etal-2010-developing,barkarson-etal-2021-mimgold} with its full tagset (553 tags in training) over the official ten folds; as in IceBERT, tokens tagged \emph{x} or \emph{e} are not scored.
\emph{NER} uses MIM-GOLD-NER \citep{ingolfsdottir-etal-2020-mimgoldner,ingolfsdottir-etal-2020-named} with IceBERT's split.
\emph{Constituency parsing} uses the hand-checked GreynirCorpus trees \citep{thorsteinsson-etal-2021-greynircorpus}: the 4,500-sentence development set for training and validation and the 500-sentence test set, with IceBERT's simplified labels.
The parser scores spans from the encoder's word vectors and decodes the best tree with CKY, in the spirit of \citet{stern-etal-2017-minimal} and \citet{kitaev-klein-2018-constituency}.
\emph{Grammatical error detection} labels each token of the Icelandic Error Corpus \citep{ingason-etal-2021-iceec} as erroneous or not, and, in a second task, with one of five error categories.
IceBERT's F-scores for this task are averaged over all classes, the no-error class included; we report that average and the F-score of the error classes alone.
\emph{Icelandic WinoGrande} \citep{snaebjarnarson-etal-2022-winogrande} uses its five folds.
The added tasks are extractive QA on NQiI and RUQuAD \citep{snaebjarnarson-einarsson-2022-natural,skarphedinsson-etal-2022-ruquad1,skarphedinsson-etal-2023-gameqa}, trained on together and scored on their pooled test questions; linguistic acceptability built ScaLA-style \citep{nielsen-2023-scandeval} from the UD Icelandic-Modern and GC treebanks; sentence-level error detection on IceEC; and dependency parsing on UD Icelandic-IcePaHC and Modern \citep{arnardottir-etal-2020-universal,zeman-etal-2025-ud216} with a biaffine parser \citep{dozat-manning-2017-deep}.

Every encoder goes through the same harness, and the learning rate is chosen on development data at seed 0 from a size-dependent grid that is extended until the choice is interior (Appendix~\ref{app:harness}).
At the chosen rate we run five seeds, once per fold for PoS and three seeds per fold for WinoGrande; the models at 10, 20 and 40 tokens per parameter in Section~\ref{sec:warm} reuse the rate of the 80-token model of the same arm and size, with three seeds.
Confidence intervals come from a hierarchical bootstrap over pretraining seeds, fine-tuning seeds and test items \citep{bouthillier-etal-2021-accounting,sellam-etal-2022-multiberts}, and differences between models are paired over the same items.
Since 45\% of the matchable IcePaHC test sentences occur in our corpus, we report IcePaHC parsing on the 58\% of its test sentences without a match; the overlaps for the other tasks are small (Appendix~\ref{app:harness}).

\section{Results}
\subsection{The IceBERT benchmarks}
\label{sec:res-main}

\input{tables/main}

Table~\ref{tab:main} gives the results.
Our 1B models are the strongest. Trained from scratch, the 1B model is significantly better than IceBERT on all eleven tasks, paired over the same test items; warm-started from Ettin, it has the best mean of all encoders on every ranked task and gains 1.3 F1 on NER, 2.2 on QA, 7.3 MCC on acceptability and 7.9 on sentence-level error detection over IceBERT.
Below 1B, our large model trained from scratch at 160 tokens per parameter, on 60B tokens against about 215B for IceBERT, is significantly better than IceBERT on PoS, NER and all three error-detection measures and not significantly different on the other tasks.
At base size, training from scratch is not enough: our base model trails IceBERT on QA and acceptability even at 160 tokens per parameter.
A warm start closes that gap. Started from mmBERT with our vocabulary and trained on 11B Icelandic tokens, the base model is not significantly worse than IceBERT on any task and is significantly better on all three error-detection measures.
The two replaced-token models trained on the Icelandic Gigaword Corpus are among the strongest below 1B on the judgement tasks (ELECTRA-base on the error categories, ConvBERT-base on binary and sentence-level error detection and second on acceptability) and do not differ significantly from each other.
The objective helps without curated text as well: mDeBERTa-v3-base \citep{he-etal-2023-debertav3}, trained with replaced-token detection on the same CC100 web text as XLM-R base, is better than XLM-R base by 12.4 MCC [7.5, 17.6] on acceptability and 6.3 [3.6, 9.8] on sentence-level error detection, so part of the Gigaword models' edge comes from the objective; whether their curated text adds to it, our comparison cannot tell.
Massively multilingual models without Icelandic-specific training are close on PoS, NER and parsing but fall far behind on acceptability: mmBERT-base scores 50.1 MCC on ScaLA against 73.6 for IceBERT.
The large warm start from Ettin, on 30B Icelandic tokens, is the best or tied best encoder below 1B on PoS tagging, NER, acceptability and both dependency treebanks, and significantly better than IceBERT on seven of the ten ranked tasks.
Icelandic WinoGrande stays near chance for every encoder up to 378M parameters that did not start from Ettin, as in the IceBERT paper; our 1B scratch model is the first clearly above chance (59.2, significantly above IceBERT's 53.8), although it saw no English and none of the test sentences occur in our corpus.
The Ettin warm starts score higher still, 67.9 at large and 78.7 at 1B, but Ettin was trained on FLAN \citep{weller-etal-2026-seq}, which contains 30,000 English WinoGrande training items with their answers \citep{wei-etal-2022-finetuned}; the base Ettin warm start saw the same data and stays at chance (51.0), but part of the two higher scores is probably supervised knowledge of the English task carried over to its translation, and we do not rank those models on this task.
One collapsed fine-tuning seed each explains the wide intervals of IceBERT-XLMR-IC3 and the large Ettin warm start on binary error detection (Appendix~\ref{app:harness}).

\subsection{Further tasks}
\label{sec:res-further}
The added tasks separate the models more than IceBERT's own.
On QA the spread is small: our 1B warm start leads with 59.1 F1, the 1B scratch model and IceBERT-large follow with 57.8 and 57.7, and IceBERT, ConvBERT-base, our two base warm starts, XLM-R large and our large model at 160 tokens per parameter are within 1.5 of IceBERT-large.
Training from scratch at base size falls behind (54.6 at 160 tokens per parameter), and so does mmBERT-base without further Icelandic training (52.6).
Acceptability and sentence-level error detection depend much more on Icelandic-specific training.
On ScaLA, every Icelandic or Nordic encoder in Table~\ref{tab:main} below 1B reaches 67--77 MCC and our 1B models 78.6 and 80.9, while XLM-R large and mDeBERTa-v3-base reach 62.4 and 62.5 and mmBERT-base 50.1.
IceBERT-XLMR-IC3, which continued XLM-R base (50.1 in our harness) on Icelandic and kept its vocabulary, recovers three quarters of that gap but stays 5.7 below IceBERT.
On sentence-level error detection ConvBERT-base is the best below 1B (53.2 MCC, 6.0 ahead of IceBERT), level with our large model at 160 tokens per parameter (52.6) and ELECTRA-base (52.0); the 1B models reach 53.5 from scratch and 55.0 warm-started.
Dependency parsing hardly separates the models: on the clean IcePaHC sentences all encoders but mmBERT-base are within 2.1 LAS of each other, and on UD Modern all but mmBERT-base and mDeBERTa-v3-base within 2.3, with the 1B warm start at the top of both.
\subsection{Warm start against scratch}
\label{sec:res-warm}

\begin{figure*}[t]
\centering
\includegraphics[width=\textwidth]{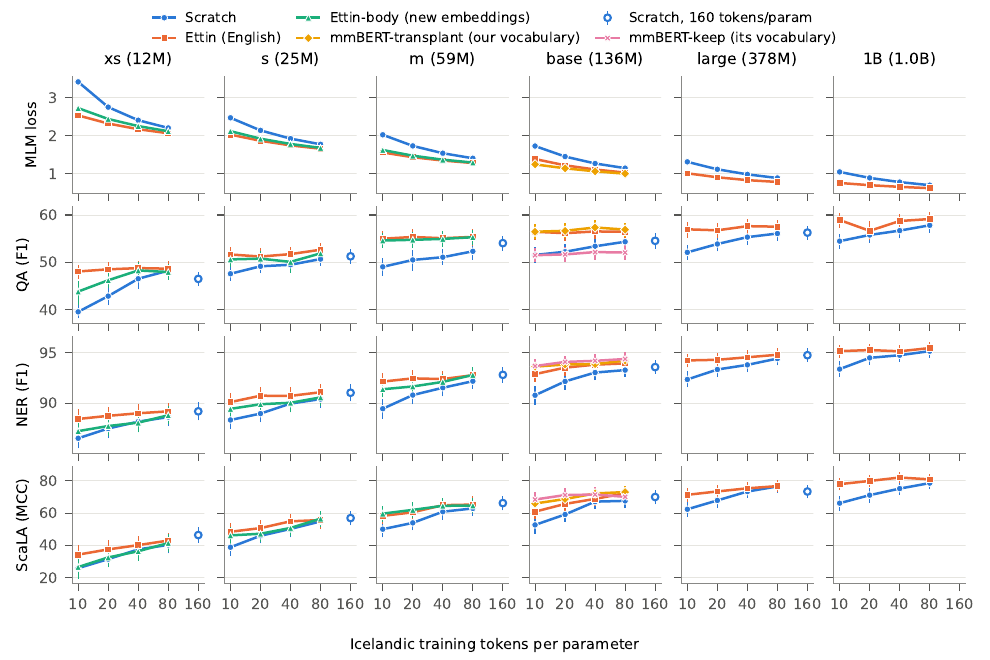}
\caption{Validation MLM loss and downstream scores against Icelandic training tokens per parameter, for each arm and size. Bars are 95\% confidence intervals (two pretraining seeds at xs, s and m; one at base, large and 1B). The open marker at 160 is the scratch model trained on twice the tokens. At large, the scratch curve is a second run of the recipe of the large model in Table~\ref{tab:main}. mmBERT-keep is left out of the loss row, as its loss is per Gemma token.}
\label{fig:curves}
\end{figure*}

Figure~\ref{fig:curves} shows the curves.
In MLM loss, a warm start from English Ettin reaches the loss of scratch training at 80 tokens per parameter with 40 or fewer at every size, and the gap at equal tokens shrinks from 0.29--0.88 nats at 10 tokens per parameter to 0.08--0.14 at 80.
Downstream, the gain is largest when Icelandic text is scarce: at 10 tokens per parameter the warm start from Ettin adds 4--9 F1 on QA and 8--12 MCC on acceptability at every size, and the one from mmBERT 5 F1 and 13 MCC.
It shrinks as the budget grows, but it does not disappear.
Against scratch training on twice the tokens, a warm start at 80 tokens per parameter is ahead on QA at every size (by 1.3 to 2.2 F1, paired 95\% intervals above zero) and level or ahead on NER and acceptability; on sentence-level error detection it is level or ahead at every size but s, where both Ettin arms are 2 MCC behind ($-2.0$ [$-3.9$, $-0.04$] for Ettin).
Outside that case a warm start halves the Icelandic data needed.
At base size the saving is larger: from Ettin, 20 tokens per parameter are not significantly worse than scratch at 160 on any of the four tasks, an eighth of the data.
From mmBERT-transplant, 10 tokens per parameter (a sixteenth) are not significantly worse on QA, NER and sentence-level error detection, but are on acceptability ($-4.1$ [$-8.6$, $-0.1$] MCC).
The acceptability intervals are wide, and at these budgets both warm starts are about 4 MCC below scratch at 160 on that task, so these savings hold for the other three tasks more firmly than for acceptability.
At large size the saving is smaller: 40 tokens per parameter from Ettin are not significantly worse than scratch at 80 or at 160 on any of the four tasks, a quarter of the data, while 10 are level with scratch at 80 on QA and NER but behind on acceptability ($-5.2$ [$-9.4$, $-1.7$] MCC) and sentence-level error detection ($-2.9$ [$-4.8$, $-0.9$]).
At 1B, 10 tokens per parameter are level with scratch at 80 on all four tasks, an eighth of the data.
Against scratch at the same budget, the warm start is ahead on all four tasks at 10 tokens per parameter at both sizes (at 1B by 4.5 F1 on QA and 11.8 MCC on acceptability), at large also at 20, and at 1B on three of the four at 20 and at 40.
Across all sizes, budgets and tasks, 104 of the 170 same-budget differences for warm starts with our vocabulary (Ettin, Ettin-body and mmBERT-transplant; Table~\ref{tab:warm-diff}) are significantly positive and none is significantly negative.
The benefit grows with model size up to m and then levels off, while the loss gap stays about the same; at 80 tokens per parameter the QA gain is $+0.4$ at xs, $+2.0$ at s, $+3.0$ at m, $+2.1$ at base, $+1.4$ at large and $+1.3$ at 1B.
Error detection and acceptability gain only at small budgets at the three smaller sizes and at large; at 1B they gain at 10, 20 and 40 tokens per parameter but not significantly at 80.
Table~\ref{tab:warm-diff} in the appendix gives every difference with its interval, and Table~\ref{tab:warm-diff160} every budget against scratch at 160.

Which part of the source model carries the transfer?
With fresh embeddings (Ettin-body), Ettin's transformer layers alone recover 62--90\% of the loss gain, and their share of the downstream gain grows with model size: at 10 tokens per parameter they give 50\%, 74\% and 93\% of the QA gain at xs, s and m and 8\%, 77\% and all of the acceptability gain (sentence-level error detection shows no trend). These shares rest on one Ettin-body seed and are ratios of two noisy differences; still, at the smallest size the transplanted embeddings carry much of the transfer, and at m the layers carry most of it.
A source that has seen Icelandic is better still: mmBERT-transplant is level with or ahead of Ettin at base on every task but PoS (where both are within 0.3 of scratch), most clearly on acceptability at small budgets (+13.3 against +8.2 MCC over scratch at 10 tokens per parameter), and it matches the loss of scratch at 80 tokens per parameter with 20.

Should the source vocabulary be kept?
mmBERT-keep, with its own 256k vocabulary, keeps embeddings that were trained on Icelandic, and at 10 tokens per parameter it gains most on acceptability (+15.7 MCC over scratch).
Paired against mmBERT-transplant, it is level on NER, PoS, acceptability and sentence-level error detection at every budget, with one exception in its favour (+2.6 MCC on error detection at 40 tokens per parameter).
On QA, however, it is 4.8 to 5.2 F1 behind at every budget, and Icelandic training does not lift it above mmBERT-base itself (51.5--52.2 against 52.6); at 80 tokens per parameter it is 2.3 F1 below scratch, the only warm-start difference in the study that favours scratch.\footnote{The pair template, offsets and segmentation are those of mmBERT's released tokenizer; the deficit is nearly as large (4.6 F1) on the questions that fit in one window, and a window that holds as much text as ours does not close it (Appendix~\ref{app:checks}).}
Keeping the vocabulary also costs more: the Gemma tokenizer splits Icelandic into 1.6 times as many tokens, and with 307M instead of 136M parameters the run over the same text took 2.8 times the compute.

\section{Discussion}
\label{sec:discussion}

\paragraph{What changed since IceBERT.}
An Icelandic encoder trained only on open data now matches or beats IceBERT on its own benchmarks, but by how much depends on the task.
PoS tagging, NER and constituency parsing were close to their ceiling in 2022 and moved by fractions of a point, while error detection moved most: our large model gains 2.5 macro F1 on binary and 4.4 on five-category token-level detection and 5.5 MCC on sentence-level detection over IceBERT (1.2, 1.7 and 4.1 over IceBERT-large, which has about the same size).
Our models differ from IceBERT in architecture, corpus, tokenizer, size and training length at once, so we cannot credit the architecture alone; what the comparison does show is that the recipe needs fewer Icelandic tokens (60B for our large model against about 215B for IceBERT).

\paragraph{A warm start is the safe default.}
IceBERT found that a multilingual warm start can give the best results on some tasks; with the warm start controlled (same shape, tokenizer, data and schedule), the finding is stronger: a warm start with our vocabulary was never significantly worse than scratch training on the same text, and its value grows as Icelandic text becomes scarce.
For a team with a fixed compute budget, a base or 1B model can be warm-started on an eighth of the text that a scratch model needs, and a large model on a quarter, without significantly lower scores on QA, NER, acceptability and sentence-level error detection, or a larger model can be trained for the same compute; on acceptability that claim is the weakest, since the intervals are wide and at base the warm starts at these budgets are about 4 MCC below scratch at 160.
Even a purely English source helps, increasingly through its transformer layers as models grow, so the lack of a source model that has seen the language is no reason to train from scratch; one that has seen it is better still on acceptability.
The vocabulary should be transplanted rather than kept: keeping mmBERT's gave no significant gain on most tasks, lost 5 F1 on QA and cost 2.8 times the compute, and IceBERT's own warm start, which kept XLM-R's vocabulary, is behind IceBERT on QA and acceptability.

\paragraph{What the benchmarks no longer measure.}
Fine-grained PoS tagging separates all models but the massively multilingual mmBERT-base and mDeBERTa-v3-base by 0.8 points, and dependency parsing by 2.3 LAS or less.
Icelandic WinoGrande stays at chance up to 378M parameters, as in 2022, and only our 1B scratch model is clearly above it without help from English WinoGrande data; the Ettin warm starts show that a source model can bring in task exposure that decontaminating the target-language corpus cannot catch.
Acceptability and error detection still separate the models; there the replaced-token objective \citep{clark-etal-2020-electra} is part of the Gigaword models' edge, and combining it with the newer architecture is the obvious next step.

\section*{Limitations}

\paragraph{Seeds.}
The warm-start curves at base size rest on one pretraining seed per arm, and each large and 1B model on one seed.
At the three smaller sizes, two pretraining seeds differed by at most 0.03 nats of validation loss, but the base intervals cover only fine-tuning seeds and test items.
At large size, the scratch curve is a second run of the recipe and seed of the large model in Table~\ref{tab:main}. The two runs reach the same validation loss within 0.004 nats and the same scores within 0.1 points on QA, NER and PoS, but the second is 3.1 MCC higher on acceptability ([$+0.02$, $+6.1$]) and 1.5 higher on sentence-level error detection, so single-run differences of this size on these two tasks may be pretraining noise that our intervals do not cover.

\paragraph{Comparison points.}
The scratch models at 160 tokens per parameter also saw the unique part of CulturaX (3.7\% more text), so they are not a pure continuation of the 80-token runs.
Fine-tuning at the shorter budgets reuses the learning rate chosen for the 80-token model of the same arm and size; a check at base size suggests that this overstates the warm-start gains at 10 tokens per parameter by one or two points (Appendix~\ref{app:checks}). Warm-started pretraining uses the scratch peak learning rate. Lower rates were no better wherever we tried them (half and a quarter at xs, half at base), but we did not try higher ones.

\paragraph{Contamination.}
We removed evaluation text from our corpus for the five tasks used first and only counted overlaps for the tasks added later; 45\% of the IcePaHC test sentences that can be matched occur in our corpus, so we report IcePaHC parsing on the clean sentences.
We could not check the training data of the source models (Ettin, mmBERT, XLM-R) or of the IceBERT models in full; it may contain test text, which would favour the warm starts and the baselines, not our scratch models. Ettin's does contain English WinoGrande training data (Section~\ref{sec:res-main}).

\paragraph{Data and scope.}
MIM-GOLD and MIM-GOLD-NER are licensed for research use, so models fine-tuned on them are not released.
We release the pretrained models but not the corpus: every source is public, but not every source states a licence that permits redistribution.
The study covers one language with 9B tokens of open text and source models of exactly matching shape; languages with less text, or without such source models, may behave differently.
We evaluate fine-tuning only; sentence embeddings and retrieval are not tested, and for long inputs we only check that the models use distant context in masked-token prediction (Appendix~\ref{app:checks}), not their performance on long-input tasks.

\makeatletter
\ifacl@anonymize\else
\section*{Acknowledgments}
This work was carried out in the TrustLLM project, funded by the European Union's Horizon Europe research and innovation programme under grant agreement No.~101135671.
We acknowledge the EuroHPC Joint Undertaking for awarding the TrustLLM project access to the EuroHPC supercomputer MareNostrum~5, hosted by the Barcelona Supercomputing Center (BSC), Spain.
Views and opinions expressed are those of the author only and do not necessarily reflect those of the European Union, which cannot be held responsible for them.
\fi
\makeatother

\bibliography{references}

\appendix

\section{Confidence intervals}
\label{app:ci}

Tables~\ref{tab:ci-ice} and~\ref{tab:ci-added} give 95\% confidence intervals for all 30 encoders: the 19 of Table~\ref{tab:main}, the small ELECTRA and ConvBERT models trained on the Gigaword Corpus \citep{gudnason-loftsson-2022-pre}, XLM-R base, mmBERT-small, our scratch models at xs, s and m, and the scratch base model from the warm-start study, which reproduces the ladder's base model. Table~\ref{tab:warm-diff} gives the paired differences behind Figure~\ref{fig:curves}, and Table~\ref{tab:warm-diff160} every budget against scratch training at 160 tokens per parameter.

\input{tables/appendix_ci}
\input{tables/appendix_warm}
\input{tables/appendix_warm160}

\section{Pretraining hyperparameters}
\label{app:hyper}

\begin{table}[t]
\centering
\small
\setlength{\tabcolsep}{3pt}
\begin{tabular}{lrrrrrr}
\toprule
Size & Heads & GeGLU & Peak LR & Batch & Warmup & WD \\
\midrule
xs & 4 & 384 & 2e-3 & 0.26M & 30M & 3e-4 \\
s & 6 & 576 & 2e-3 & 0.26M & 61M & 3e-4 \\
m & 8 & 768 & 1.5e-3 & 0.52M & 143M & 3e-4 \\
base & 12 & 1152 & 8e-4 & 1.05M & 327M & 1e-5 \\
large & 16 & 2624 & 4e-4 & 1.05M & 907M & 1e-5 \\
1B & 28 & 3840 & 3e-4 & 2.10M & 2.40B & 5e-5 \\
\bottomrule
\end{tabular}
\caption{Pretraining hyperparameters per size (layers and width are in Table~\ref{tab:sizes}). GeGLU: inner width of the feed-forward layer. Batch and warmup in tokens; warmup is 3\% of the 80-token budget. WD: decoupled weight decay.}
\label{tab:hyper}
\end{table}

Table~\ref{tab:hyper} lists what differs between sizes. All models use AdamW ($\beta_1 = 0.9$, $\beta_2 = 0.98$, $\epsilon = 10^{-6}$), gradient clipping at 1.0 and bf16 autocast; tied input and output embeddings; no biases; a 128-token sliding window in two of every three layers and global attention in the third; and rotary embeddings with base 160k in the global and 10k in the local layers (the warm starts keep their source models' 160k in both). Masking follows BERT's 80/10/10 split at 30\%, and at 15\% during the decay. The first 90\% of the tokens use 1,024-token sequences, the rest 8,192.

\section{Data pipeline}
\label{app:data}

\begin{table}[t]
\centering
\small
\begin{tabular}{lrrr}
\toprule
Source & Docs in & Shared & Tokens \\
\midrule
IGC-2024 (open) & 2.15M & 22\% & 2.01B \\
FinePDFs$^\ast$ & 0.36M & 8\% & 1.86B \\
FineWeb-2$^\ast$ & 3.05M & 84\% & 1.61B \\
HPLT 3.0$^\ast$ & 4.30M & 65\% & 1.38B \\
MaCoCu$^\ast$ & 2.42M & 34\% & 1.17B \\
MADLAD-400$^\ast$ & 1.56M & 63\% & 0.56B \\
IC3-v2 & 3.86M & 69\% & 0.36B \\
Dynaword (non-IGC)$^\ast$ & 0.11M & 53\% & 0.06B \\
Wikipedia$^\ast$ & 0.06M & 76\% & $<$0.01B \\
\midrule
Total & 17.86M & & 9.02B \\
\midrule
CulturaX (160 only) & 2.37M & 77\% & 0.34B \\
\bottomrule
\end{tabular}
\caption{Pretraining sources. \emph{Shared}: share of a source's documents with a near-duplicate in another source. \emph{Tokens}: training tokens after filtering, deduplication in the listed priority order (IGC first) and decontamination, in our 32k tokenizer. $^\ast$Used in the 8,192-token phase (IGC and IC3-v2 do not keep paragraph order). CulturaX was added for the 160 tokens per parameter models only, deduplicated last against the nine sources, which stay unchanged.}
\label{tab:data}
\end{table}

The filter drops a document at the first rule it fails: fewer than 25 words (10 for Wikipedia and Dynaword); an Icelandic function-word score below 0.10, or a closer match to Faroese, Danish and Norwegian, Swedish or English function words; fewer than 60\% alphabetic characters, or no Icelandic-specific letter (á, ð, é, í, ó, ú, ý, þ, æ, ö) in a document of over 100 words; duplicated lines holding more than 30\% of the characters, or a single word 3-gram covering more than 20\% of the text; and more than 30\% of lines shorter than four words that end without punctuation. Lines matching cookie, consent and navigation patterns are removed first. Deduplication removes exact copies, then near-duplicates by MinHash over word 5-grams (112 hash functions, 14 bands of 8 rows), keeping candidate pairs whose estimated Jaccard similarity is at least 0.75 and clustering them by union-find; each cluster keeps the copy from the highest-priority source in the order of Table~\ref{tab:data}, then the longest. Decontamination matches lower-cased word sequences as described in Section~\ref{sec:data}.

\section{Fine-tuning and evaluation details}
\label{app:harness}

Every encoder goes through the same harness: a linear head (or the parser) on the final layer, with the first sub-token representing each word in token tasks and mean pooling in sentence tasks; AdamW with batch size 32; and the test score of the epoch that is best on development data.
QA uses 384-token windows with a stride of 128.
The learning rate is chosen on development data at seed 0 from a three-value grid that depends on model size; when the chosen rate is at an edge of the grid, two more values (twice and four times, or a half and a quarter of the edge) are added until the choice is interior.
This rule matters: on the new tasks the initial grids chose an edge in 132 of 161 cases.
We exempt WinoGrande, where most models are near chance and a development set of about 85 items cannot rank learning rates.
Re-tuning the shorter-budget models at base size and 10 tokens per parameter lowers the warm-start gain by one or two points but leaves it significant (Appendix~\ref{app:checks}).
Fine-tuning is mostly stable at the chosen learning rates. Two runs of binary error detection are not: a seed of IceBERT-XLMR-IC3 predicts no errors at all, and a seed of the large Ettin warm start falls to 62.3; without them, the two models score 68.0 and 71.2. Collapsed runs also occur at high learning rates during the grid search, where development data rejects them.
Confidence intervals come from a hierarchical bootstrap \citep{bouthillier-etal-2021-accounting,sellam-etal-2022-multiberts}: each of 1,000 draws resamples pretraining seeds (where there are several), then fine-tuning seeds within each model, then test items, and recomputes the corpus-level metric (F1, MCC, accuracy or LAS) on the resampled items.
Fold tasks pool their test sets.
Differences between models are paired over the same item draws.
Our scores for IceBERT are close to those reported in its paper on constituency parsing, error detection and WinoGrande, but differ on PoS tagging (97.8 against 98.3) and NER (94.1 against 91.4), so we compare models only within our harness.

Text from the five tasks we used first was removed from the pretraining data (Section~\ref{sec:data}).
For the tasks added later we counted overlaps instead: none of the GreynirCorpus or WinoGrande test sentences occur in our corpus, and only 0.2\% of the MIM-GOLD PoS and 0.1\% of the error-detection test sentences, but 2,234 of the 5,003 IcePaHC test sentences long enough to match (45\%) do, since the treebank consists of historical texts such as the sagas that circulate widely online.
We therefore report IcePaHC parsing on the 2,968 of its 5,157 test sentences (58\%) that contain no match, and give the full test set in Appendix~\ref{app:ci}.

The initial learning-rate grid depends on encoder size: $\{1, 2, 4\} \times 10^{-4}$ up to 40M parameters, $\{3, 6, 10\} \times 10^{-5}$ up to 200M, $\{1, 2, 4\} \times 10^{-5}$ up to 500M, and $\{0.7, 1.5, 3\} \times 10^{-5}$ above; for the five original tasks our large and 1B models start from per-task grids of three to seven values between $10^{-5}$ and $1.6 \times 10^{-4}$. Training runs for 3 epochs on NER, MIM-GOLD PoS, token-level error detection, sentence-level error detection and QA; 5 on UD PoS, acceptability and dependency parsing; 10 on constituency parsing and WinoGrande. Inputs are truncated at 512 tokens for token classification and 256 for sentence and WinoGrande inputs. All runs use linear warmup over 10\% of the steps, then linear decay, weight decay 0.01, gradient clipping at 1.0 and bf16 autocast; a run whose loss becomes non-finite in bf16 is repeated in fp32 and flagged (this affected ScandiBERT). mDeBERTa-v3-base is stored in fp16; like every other model, it is fine-tuned from fp32 weights.

\section{Additional checks}
\label{app:checks}

\paragraph{Learning rates at short budgets.}
The models at 10, 20 and 40 tokens per parameter reuse the fine-tuning learning rate of the 80-token model of their arm and size.
To test this shortcut where it matters most, we re-tuned the base Scratch and Ettin models at 10 tokens per parameter on QA and acceptability with the full rule of Section~\ref{sec:eval} (the 80-token model's grid at seed 0, extended until the choice is interior, then three seeds).
Three of the four rates change, and re-tuning helps Scratch on acceptability most (Table~\ref{tab:check-lr}).
The warm-start gain shrinks from 4.9 to 4.2 F1 on QA and from 8.2 to 6.5 MCC on acceptability, and stays significant on both.
The reused rates therefore overstate the gains at the smallest budget by one or two points, but do not create them.

\paragraph{Pretraining learning rate.}
Every warm-started arm is pretrained at the scratch peak learning rate, which the pilot of Section~\ref{sec:warm} supported only at xs.
To test whether a warm start at a larger size needs a gentler rate, we trained Ettin and mmBERT-keep at base size straight to 10 tokens per parameter at half the rate, with the schedule otherwise unchanged, and fine-tuned them on four tasks at the rates of the study's 10-token models (Table~\ref{tab:check-pretrain-lr}).
For Ettin, half the rate ends 0.17 nats higher in masked-token loss and is no better on any task: QA and NER barely change, while acceptability falls by 6.6 MCC and sentence-level error detection by 2.6, so on those two tasks the gain over scratch training shrinks to 1.7 and 1.5 and is no longer significant.
For mmBERT-keep, half the rate changes the loss by 0.01 nats and no task significantly.
A lower rate therefore helps neither warm start, so the scratch rate does not understate the gains; the Ettin gains on acceptability and error detection at this budget do depend on the rate.

\paragraph{Long inputs.}
To check that the 8,192-token phase worked, we score masked tokens in the last 1,024 positions of the 54 held-out documents of at least 8,192 tokens from the sources used in that phase, giving the model the last 1k, 2k, 4k or all 8k tokens (15\% of positions masked, the same positions for every model).
For every one of the 25 models with our vocabulary at 80 or 160 tokens per parameter, the full context lowers the loss, by 0.04 to 0.09 nats (Table~\ref{tab:check-longctx}); most of the gain is already there with 4k tokens.
After the first 512 positions, loss in the full input also falls with position.
The models therefore use context well beyond the 1,024 tokens of most of their training; this does not test long-input tasks.

\paragraph{mmBERT-keep on QA.}
The Gemma tokenizer needs 1.6 times as many tokens for the same text, so mmBERT-keep's 384-token QA windows hold less of each passage.
With 614-token windows (stride 204), which hold the same text as 384 of our tokens, mmBERT-keep at 80 tokens per parameter reaches 52.7 F1 against 52.1, and mmBERT-transplant gains the same 0.6 from the wider window (57.5 against 56.9; five seeds each, neither change significant).
mmBERT-keep with the wider window is still 4.2 F1 [3.3, 5.1] behind mmBERT-transplant with the usual one, against 4.8 with equal windows, and exact match shows the same gap (29.6 against 35.0).
The window explains at most a small part of the deficit.

\input{tables/appendix_checks}

\end{document}

%% file: tables/main.tex
\begin{table*}[t]
\centering
\small
\setlength{\tabcolsep}{3.3pt}
\begin{tabular}{llrrrrrrrrrrr}
\toprule
Model & Size & PoS & NER & Const. & GED & GED-5 & WinoG. & QA & ScaLA & IceEC & LAS-H & LAS-M \\
\midrule
\multicolumn{13}{l}{\emph{Icelandic and Nordic baselines}} \\
IceBERT & 163M & 97.8 & 94.1 & 90.5 & 68.5 & 32.9 & 53.8 & 56.9 & 73.6 & 47.1 & 86.6 & 79.2 \\
IceBERT-large & 406M & 97.8 & 94.3 & 91.2 & 69.8 & 35.6 & 54.1 & 57.7 & 73.2 & 48.5 & 86.9 & 79.0 \\
IceBERT-XLMR-IC3 & 279M & 97.6 & 93.9 & 90.2 & 64.2$^\ddagger$ & 30.3 & 50.8 & 55.0 & 67.9 & 46.7 & 86.4 & 78.5 \\
ELECTRA-base (IGC) & 110M & 97.5 & 93.5 & 90.1 & 71.1 & 38.4 & 52.3 & 56.0 & 75.1 & 52.0 & 85.6 & 78.8 \\
ConvBERT-base (IGC) & 107M & 97.6 & 94.1 & 90.5 & 71.3 & 36.5 & 52.4 & 56.4 & 76.3 & 53.2 & 86.0 & 79.3 \\
ScandiBERT & 125M & 97.6 & 93.5 & 90.2 & 68.2 & 33.4 & 53.9 & 54.3 & 68.1 & 46.7 & 86.4 & 78.6 \\
\midrule
\multicolumn{13}{l}{\emph{Multilingual baselines}} \\
XLM-R large & 560M & 97.3 & 93.5 & 90.4 & 68.6 & 31.3 & 51.7 & 56.5 & 62.4 & 45.7 & 86.7 & 78.4 \\
mDeBERTa-v3-base & 278M & 96.8 & 92.6 & 89.9 & 67.5 & 31.1 & 52.6 & 55.0 & 62.5 & 47.3 & 85.7 & 77.6 \\
mmBERT-base & 307M & 96.8 & 92.3 & 89.1 & 67.1 & 30.7 & 50.7 & 52.6 & 50.1 & 42.4 & 85.0 & 76.8 \\
\midrule
\multicolumn{13}{l}{\emph{Ours, from scratch}} \\
base, 80 & 136M & 97.4 & 93.3 & 90.1 & 69.1 & 35.5 & 50.1 & 54.3 & 67.5 & 47.4 & 85.5 & 78.1 \\
base, 160 & 136M & 97.6 & 93.6 & 89.8 & 69.4 & 35.1 & 50.1 & 54.6 & 70.0 & 48.2 & 85.8 & 78.4 \\
large, 80 & 378M & 97.8 & 94.5 & 90.2 & 70.6 & 36.9 & 48.9 & 56.0 & 73.4 & 50.0 & 86.2 & 79.0 \\
large, 160 & 378M & 97.9 & 94.7 & 90.9 & 71.0 & 37.4 & 51.6 & 56.3 & 73.4 & 52.6 & 86.7 & 78.9 \\
1B, 80 & 1.0B & 98.0 & 95.1 & 91.0 & 71.7 & 39.7 & \textbf{59.2} & 57.8 & 78.6 & 53.5 & 87.0 & 79.8 \\
\midrule
\multicolumn{13}{l}{\emph{Ours, warm start (80 tokens/param)}} \\
base, Ettin & 136M & 97.7 & 93.9 & 90.5 & 69.4 & 34.9 & 51.0 & 56.5 & 72.3 & 48.4 & 86.0 & 79.0 \\
base, mmBERT-transplant & 136M & 97.8 & 94.2 & 90.4 & 69.6 & 36.8 & 53.0 & 56.9 & 73.0 & 49.9 & 86.5 & 79.2 \\
base, mmBERT-keep & 307M & 97.8 & 94.4 & 90.5 & 69.5 & 36.2 & 50.1 & 52.1 & 70.1 & 49.6 & 86.0 & 78.6 \\
large, Ettin & 378M & 97.9 & 94.8 & 91.1 & 69.4$^\ddagger$ & 37.3 & 67.9$^\dagger$ & 57.5 & 76.7 & 52.5 & 87.1 & 79.9 \\
1B, Ettin & 1.0B & \textbf{98.1} & \textbf{95.4} & \textbf{91.4} & \textbf{72.1} & \textbf{40.3} & 78.7$^\dagger$ & \textbf{59.1} & \textbf{80.9} & \textbf{55.0} & \textbf{87.5} & \textbf{80.3} \\
\bottomrule
\end{tabular}
\caption{Fine-tuning results ($\times$100), means over five fine-tuning seeds (PoS: the ten official MIM-GOLD folds; WinoGrande: five folds $\times$ three seeds). IceBERT's tasks: PoS accuracy, NER entity F1, constituency F1 (GreynirCorpus), token-level error detection macro F1 (binary and five categories), WinoGrande accuracy. Added: QA token F1 (NQiI + RUQuAD), ScaLA and sentence-level IceEC MCC, LAS on UD IcePaHC (the 58\% of test sentences not in our pretraining data) and UD Modern. Warm starts: from Ettin, or from mmBERT with our vocabulary transplanted (mmBERT-transplant) or with its own vocabulary kept (mmBERT-keep). Bold: best mean per column. $^\dagger$Not ranked: the source model was trained on English WinoGrande training data (Section~\ref{sec:res-main}). $^\ddagger$One of five fine-tuning seeds collapsed (Section~\ref{sec:res-main}). Appendix~\ref{app:ci} gives 95\% confidence intervals for these and 11 further encoders.}
\label{tab:main}
\end{table*}

%% file: tables/appendix_ci.tex
\begin{table*}[t]
\centering
\small
\setlength{\tabcolsep}{2.4pt}
\begin{tabular}{lcccccccc}
\toprule
Model & PoS & NER & Const. & GED & GED$_e$ & GED-5 & GED-5$_e$ & WinoG. \\
\midrule
IceBERT & 97.8$\pm$0.3 & 94.1$\pm$0.7 & 90.5$\pm$0.7 & 68.5$\pm$1.0 & 39.1$\pm$2.1 & 32.9$\pm$1.3 & 34.2$\pm$2.3 & 53.8$\pm$2.7 \\
IceBERT-large & 97.8$\pm$0.3 & 94.3$\pm$0.7 & 91.2$\pm$0.7 & 69.8$\pm$1.1 & 41.6$\pm$2.0 & 35.6$\pm$2.7 & 35.7$\pm$2.3 & 54.1$\pm$2.9 \\
IceBERT-XLMR-IC3 & 97.6$\pm$0.3 & 93.9$\pm$0.7 & 90.2$\pm$0.8 & 64.2$\pm$7.8$^\ddagger$ & 30.4$\pm$15.5 & 30.3$\pm$1.3 & 31.6$\pm$2.2 & 50.8$\pm$2.5 \\
ELECTRA-base (IGC) & 97.5$\pm$0.3 & 93.5$\pm$0.7 & 90.1$\pm$0.7 & 71.1$\pm$1.0 & 44.3$\pm$2.0 & 38.4$\pm$3.5 & 38.0$\pm$2.2 & 52.3$\pm$3.0 \\
ConvBERT-base (IGC) & 97.6$\pm$0.3 & 94.1$\pm$0.7 & 90.5$\pm$0.6 & 71.3$\pm$1.0 & 44.5$\pm$2.0 & 36.5$\pm$2.3 & 38.6$\pm$2.3 & 52.4$\pm$2.7 \\
ScandiBERT & 97.6$\pm$0.3 & 93.5$\pm$0.7 & 90.2$\pm$0.7 & 68.2$\pm$1.0 & 38.5$\pm$2.1 & 33.4$\pm$2.6 & 33.7$\pm$2.3 & 53.9$\pm$2.6 \\
ELECTRA-small (IGC) & 96.4$\pm$0.3 & 90.3$\pm$0.8 & 89.0$\pm$0.8 & 67.9$\pm$1.1 & 37.9$\pm$2.0 & 31.2$\pm$1.4 & 31.2$\pm$2.2 & 51.7$\pm$3.4 \\
ConvBERT-small (IGC) & 96.6$\pm$0.3 & 92.3$\pm$0.8 & 89.5$\pm$0.8 & 68.7$\pm$1.0 & 39.4$\pm$2.2 & 34.0$\pm$2.6 & 34.2$\pm$2.1 & 54.0$\pm$2.7 \\
\midrule
XLM-R large & 97.3$\pm$0.3 & 93.5$\pm$0.7 & 90.4$\pm$0.7 & 68.6$\pm$1.1 & 39.2$\pm$2.1 & 31.3$\pm$1.8 & 32.7$\pm$2.4 & 51.7$\pm$2.6 \\
mDeBERTa-v3-base & 96.8$\pm$0.3 & 92.6$\pm$0.7 & 89.9$\pm$0.8 & 67.5$\pm$1.0 & 37.1$\pm$1.9 & 31.1$\pm$2.2 & 32.2$\pm$2.3 & 52.6$\pm$2.6 \\
mmBERT-base & 96.8$\pm$0.4 & 92.3$\pm$0.8 & 89.1$\pm$0.8 & 67.1$\pm$1.1 & 36.3$\pm$2.1 & 30.7$\pm$1.8 & 30.5$\pm$2.2 & 50.7$\pm$2.4 \\
XLM-R base & 96.5$\pm$0.3 & 91.3$\pm$0.8 & 89.9$\pm$0.7 & 65.3$\pm$1.0 & 32.6$\pm$2.0 & 27.7$\pm$1.1 & 27.9$\pm$2.5 & 53.0$\pm$4.0 \\
mmBERT-small & 96.3$\pm$0.4 & 90.9$\pm$0.7 & 88.2$\pm$0.8 & 65.4$\pm$1.1 & 33.0$\pm$2.4 & 28.3$\pm$1.2 & 27.8$\pm$2.1 & 51.4$\pm$2.7 \\
\midrule
xs, 80 & 95.6$\pm$0.4 & 88.7$\pm$1.0 & 86.4$\pm$0.8 & 65.3$\pm$1.0 & 32.9$\pm$1.9 & 28.1$\pm$1.2 & 26.9$\pm$2.2 & 50.8$\pm$2.4 \\
xs, 160 & 95.8$\pm$0.4 & 89.2$\pm$0.9 & 86.9$\pm$0.8 & 65.8$\pm$1.0 & 33.8$\pm$2.0 & 28.7$\pm$1.4 & 28.2$\pm$2.3 & 50.8$\pm$2.4 \\
s, 80 & 96.5$\pm$0.4 & 90.3$\pm$0.9 & 88.0$\pm$0.8 & 67.0$\pm$1.0 & 36.3$\pm$1.9 & 31.3$\pm$3.0 & 29.5$\pm$2.2 & 50.5$\pm$2.3 \\
s, 160 & 96.7$\pm$0.4 & 91.0$\pm$0.8 & 88.3$\pm$0.8 & 67.3$\pm$1.0 & 36.8$\pm$2.0 & 30.8$\pm$2.7 & 29.9$\pm$2.2 & 49.9$\pm$2.6 \\
m, 80 & 97.2$\pm$0.4 & 91.9$\pm$0.8 & 89.3$\pm$0.8 & 68.2$\pm$1.0 & 38.5$\pm$1.8 & 33.5$\pm$3.7 & 32.1$\pm$2.2 & 49.7$\pm$2.6 \\
m, 160 & 97.3$\pm$0.4 & 92.8$\pm$0.8 & 89.7$\pm$0.7 & 68.8$\pm$1.0 & 39.7$\pm$1.9 & 33.7$\pm$3.4 & 33.0$\pm$2.2 & 49.8$\pm$2.4 \\
base, 80 & 97.4$\pm$0.4 & 93.3$\pm$0.7 & 90.1$\pm$0.7 & 69.1$\pm$0.9 & 40.3$\pm$1.9 & 35.5$\pm$3.8 & 33.6$\pm$2.2 & 50.1$\pm$2.5 \\
base, 80 (study run) & 97.5$\pm$0.4 & 93.3$\pm$0.7 & 89.9$\pm$0.8 & 69.2$\pm$1.0 & 40.6$\pm$1.8 & 35.7$\pm$4.1 & 34.3$\pm$2.2 & 50.4$\pm$2.7 \\
base, 160 & 97.6$\pm$0.4 & 93.6$\pm$0.7 & 89.8$\pm$0.9 & 69.4$\pm$1.0 & 41.0$\pm$2.0 & 35.1$\pm$3.4 & 33.6$\pm$2.1 & 50.1$\pm$2.2 \\
large, 80 & 97.8$\pm$0.4 & 94.5$\pm$0.7 & 90.2$\pm$0.7 & 70.6$\pm$1.0 & 43.2$\pm$1.9 & 36.9$\pm$3.7 & 36.1$\pm$2.3 & 48.9$\pm$2.5 \\
large, 160 & 97.9$\pm$0.4 & 94.7$\pm$0.6 & 90.9$\pm$0.7 & 71.0$\pm$1.0 & 44.1$\pm$2.0 & 37.4$\pm$3.7 & 37.0$\pm$2.3 & 51.6$\pm$2.6 \\
1B, 80 & 98.0$\pm$0.4 & 95.1$\pm$0.7 & 91.0$\pm$0.7 & 71.7$\pm$1.0 & 45.3$\pm$2.1 & 39.7$\pm$4.5 & 38.8$\pm$2.2 & 59.2$\pm$4.1 \\
\midrule
base, Ettin & 97.7$\pm$0.4 & 93.9$\pm$0.7 & 90.5$\pm$0.7 & 69.4$\pm$1.2 & 40.9$\pm$2.1 & 34.9$\pm$3.2 & 34.7$\pm$2.2 & 51.0$\pm$2.9 \\
base, mmBERT-transplant & 97.8$\pm$0.4 & 94.2$\pm$0.7 & 90.4$\pm$0.7 & 69.6$\pm$1.1 & 41.2$\pm$2.4 & 36.8$\pm$3.5 & 36.6$\pm$2.4 & 53.0$\pm$3.6 \\
base, mmBERT-keep & 97.8$\pm$0.4 & 94.4$\pm$0.7 & 90.5$\pm$0.7 & 69.5$\pm$1.0 & 41.1$\pm$1.9 & 36.2$\pm$3.5 & 35.1$\pm$2.2 & 50.1$\pm$2.5 \\
large, Ettin & 97.9$\pm$0.4 & 94.8$\pm$0.6 & 91.1$\pm$0.7 & 69.4$\pm$3.5$^\ddagger$ & 40.8$\pm$7.1 & 37.3$\pm$3.2 & 37.3$\pm$2.2 & 67.9$\pm$4.5$^\dagger$ \\
1B, Ettin & 98.1$\pm$0.4 & 95.4$\pm$0.5 & 91.4$\pm$0.7 & 72.1$\pm$1.0 & 46.3$\pm$2.0 & 40.3$\pm$4.2 & 40.1$\pm$2.3 & 78.7$\pm$2.4$^\dagger$ \\
\bottomrule
\end{tabular}
\caption{Means and 95\% confidence intervals ($\pm$ the larger half-width) for IceBERT's tasks, all 30 encoders; hierarchical bootstrap over fine-tuning seeds and test items (1,000 draws). Groups as in Table~\ref{tab:main}; the scratch models are named by size and tokens per parameter. GED$_e$ and GED-5$_e$: F1 of the error classes alone (micro-averaged over the error tokens); GED and GED-5 average F1 over all classes, the no-error class included, as IceBERT did. $^\dagger$, $^\ddagger$: as in Table~\ref{tab:main}.}
\label{tab:ci-ice}
\end{table*}

\begin{table*}[t]
\centering
\small
\setlength{\tabcolsep}{2.4pt}
\begin{tabular}{lcccccc}
\toprule
Model & QA & ScaLA & IceEC & LAS-H & LAS-H (all) & LAS-M \\
\midrule
IceBERT & 56.9$\pm$1.5 & 73.6$\pm$3.8 & 47.1$\pm$2.3 & 86.6$\pm$0.4 & 87.2$\pm$0.3 & 79.2$\pm$1.0 \\
IceBERT-large & 57.7$\pm$1.6 & 73.2$\pm$4.0 & 48.5$\pm$2.3 & 86.9$\pm$0.5 & 87.4$\pm$0.3 & 79.0$\pm$0.9 \\
IceBERT-XLMR-IC3 & 55.0$\pm$1.5 & 67.9$\pm$4.0 & 46.7$\pm$2.5 & 86.4$\pm$0.5 & 86.9$\pm$0.3 & 78.5$\pm$1.0 \\
ELECTRA-base (IGC) & 56.0$\pm$1.5 & 75.1$\pm$3.7 & 52.0$\pm$2.4 & 85.6$\pm$0.5 & 86.2$\pm$0.3 & 78.8$\pm$1.0 \\
ConvBERT-base (IGC) & 56.4$\pm$1.4 & 76.3$\pm$3.9 & 53.2$\pm$2.4 & 86.0$\pm$0.4 & 86.5$\pm$0.3 & 79.3$\pm$1.0 \\
ScandiBERT & 54.3$\pm$2.7 & 68.1$\pm$4.5 & 46.7$\pm$2.6 & 86.4$\pm$0.4 & 86.8$\pm$0.3 & 78.6$\pm$0.9 \\
ELECTRA-small (IGC) & 50.1$\pm$1.5 & 72.5$\pm$3.9 & 46.7$\pm$2.4 & 83.5$\pm$0.6 & 84.1$\pm$0.5 & 76.5$\pm$1.0 \\
ConvBERT-small (IGC) & 54.0$\pm$1.5 & 72.8$\pm$3.9 & 46.5$\pm$2.5 & 84.3$\pm$0.5 & 85.0$\pm$0.4 & 78.3$\pm$1.0 \\
\midrule
XLM-R large & 56.5$\pm$1.5 & 62.4$\pm$4.3 & 45.7$\pm$2.2 & 86.7$\pm$0.4 & 87.1$\pm$0.3 & 78.4$\pm$1.0 \\
mDeBERTa-v3-base & 55.0$\pm$1.4 & 62.5$\pm$4.2 & 47.3$\pm$2.6 & 85.7$\pm$0.5 & 86.2$\pm$0.3 & 77.6$\pm$1.0 \\
mmBERT-base & 52.6$\pm$1.5 & 50.1$\pm$4.9 & 42.4$\pm$2.4 & 85.0$\pm$0.4 & 85.7$\pm$0.3 & 76.8$\pm$1.0 \\
XLM-R base & 54.1$\pm$1.5 & 50.1$\pm$5.3 & 41.0$\pm$3.5 & 85.5$\pm$0.4 & 86.0$\pm$0.3 & 76.8$\pm$1.0 \\
mmBERT-small & 50.7$\pm$1.4 & 37.5$\pm$5.0 & 40.8$\pm$2.4 & 84.4$\pm$0.5 & 85.0$\pm$0.4 & 75.0$\pm$1.0 \\
\midrule
xs, 80 & 48.2$\pm$1.5 & 43.2$\pm$5.5 & 38.6$\pm$2.7 & 80.6$\pm$0.5 & 81.7$\pm$0.4 & 73.8$\pm$1.0 \\
xs, 160 & 46.5$\pm$1.5 & 46.4$\pm$5.0 & 39.6$\pm$2.5 & 81.2$\pm$0.5 & 82.2$\pm$0.4 & 74.1$\pm$1.0 \\
s, 80 & 50.3$\pm$2.0 & 51.2$\pm$4.8 & 41.2$\pm$2.5 & 82.8$\pm$0.5 & 83.7$\pm$0.4 & 75.9$\pm$0.9 \\
s, 160 & 51.3$\pm$1.5 & 56.9$\pm$4.6 & 43.5$\pm$2.5 & 83.3$\pm$0.5 & 84.1$\pm$0.4 & 76.6$\pm$1.0 \\
m, 80 & 52.7$\pm$1.5 & 63.2$\pm$4.6 & 45.2$\pm$2.5 & 84.5$\pm$0.4 & 85.1$\pm$0.4 & 77.0$\pm$1.0 \\
m, 160 & 54.1$\pm$1.5 & 66.2$\pm$4.1 & 46.4$\pm$2.4 & 85.2$\pm$0.5 & 85.8$\pm$0.4 & 78.0$\pm$1.0 \\
base, 80 & 54.3$\pm$1.6 & 67.5$\pm$4.2 & 47.4$\pm$2.5 & 85.5$\pm$0.5 & 86.0$\pm$0.4 & 78.1$\pm$1.0 \\
base, 80 (study run) & 54.4$\pm$1.5 & 67.6$\pm$4.0 & 47.3$\pm$2.7 & 85.4$\pm$0.4 & 86.0$\pm$0.3 & 78.2$\pm$0.9 \\
base, 160 & 54.6$\pm$1.6 & 70.0$\pm$4.0 & 48.2$\pm$2.5 & 85.8$\pm$0.4 & 86.3$\pm$0.3 & 78.4$\pm$1.0 \\
large, 80 & 56.0$\pm$1.5 & 73.4$\pm$3.9 & 50.0$\pm$2.4 & 86.2$\pm$0.5 & 86.8$\pm$0.3 & 79.0$\pm$1.0 \\
large, 160 & 56.3$\pm$1.5 & 73.4$\pm$4.2 & 52.6$\pm$2.3 & 86.7$\pm$0.4 & 87.2$\pm$0.4 & 78.9$\pm$0.9 \\
1B, 80 & 57.8$\pm$1.5 & 78.6$\pm$3.5 & 53.5$\pm$2.4 & 87.0$\pm$0.4 & 87.6$\pm$0.3 & 79.8$\pm$0.9 \\
\midrule
base, Ettin & 56.5$\pm$1.5 & 72.3$\pm$3.8 & 48.4$\pm$2.5 & 86.0$\pm$0.4 & 86.6$\pm$0.3 & 79.0$\pm$1.0 \\
base, mmBERT-transplant & 56.9$\pm$1.5 & 73.0$\pm$3.7 & 49.9$\pm$2.4 & 86.5$\pm$0.4 & 87.1$\pm$0.3 & 79.2$\pm$1.0 \\
base, mmBERT-keep & 52.1$\pm$1.5 & 70.1$\pm$4.5 & 49.6$\pm$2.4 & 86.0$\pm$0.4 & 86.6$\pm$0.3 & 78.6$\pm$1.0 \\
large, Ettin & 57.5$\pm$1.4 & 76.7$\pm$3.6 & 52.5$\pm$2.4 & 87.1$\pm$0.4 & 87.5$\pm$0.3 & 79.9$\pm$1.0 \\
1B, Ettin & 59.1$\pm$1.6 & 80.9$\pm$3.6 & 55.0$\pm$2.2 & 87.5$\pm$0.4 & 88.0$\pm$0.3 & 80.3$\pm$0.9 \\
\bottomrule
\end{tabular}
\caption{Means and 95\% confidence intervals ($\pm$ the larger half-width) for added tasks, all 30 encoders; hierarchical bootstrap over fine-tuning seeds and test items (1,000 draws). Groups as in Table~\ref{tab:main}; the scratch models are named by size and tokens per parameter. LAS-H: the clean 58\% of the IcePaHC test sentences; LAS-H (all): the full test set. $^\dagger$, $^\ddagger$: as in Table~\ref{tab:main}.}
\label{tab:ci-added}
\end{table*}

%% file: tables/appendix_warm.tex
\begin{table*}[t]
\centering
\small
\setlength{\tabcolsep}{1.7pt}
\resizebox{\textwidth}{!}{%
\begin{tabular}{lllccccc}
\toprule
& & & \multicolumn{4}{c}{vs.\ scratch, same budget} & vs.\ scratch \\
\cmidrule(lr){4-7}
Task & Size & Arm & 10 & 20 & 40 & 80 & 80 vs.\ 160 \\
\midrule
QA & xs & Ettin & \textbf{+8.5 [+7.3, +9.8]} & \textbf{+5.7 [+3.9, +7.4]} & \textbf{+2.2 [+0.4, +4.4]} & +0.4 [-1.1, +1.8] & \textbf{+2.2 [+0.8, +3.5]} \\
 & xs & Ettin-body & \textbf{+4.3 [+1.7, +6.4]} & \textbf{+3.3 [+1.5, +5.0]} & +1.8 [-0.2, +3.9] & -0.3 [-1.6, +1.0] & \textbf{+1.4 [+0.2, +2.8]} \\
 & s & Ettin & \textbf{+4.1 [+3.0, +5.1]} & \textbf{+2.1 [+0.9, +3.2]} & \textbf{+2.2 [+1.0, +3.6]} & \textbf{+2.0 [+1.0, +3.1]} & \textbf{+1.4 [+0.4, +2.3]} \\
 & s & Ettin-body & \textbf{+3.0 [+2.0, +4.1]} & \textbf{+1.6 [+0.4, +2.6]} & +0.6 [-1.6, +2.5] & \textbf{+1.2 [+0.02, +2.5]} & +0.6 [-0.5, +1.9] \\
 & m & Ettin & \textbf{+5.9 [+4.3, +7.7]} & \textbf{+4.8 [+2.7, +7.1]} & \textbf{+4.0 [+2.5, +5.4]} & \textbf{+3.0 [+1.5, +4.5]} & \textbf{+1.3 [+0.1, +2.6]} \\
 & m & Ettin-body & \textbf{+5.5 [+3.9, +7.3]} & \textbf{+4.2 [+2.3, +6.3]} & \textbf{+3.9 [+2.5, +5.0]} & \textbf{+3.0 [+1.7, +4.3]} & \textbf{+1.3 [+0.3, +2.2]} \\
 & base & Ettin & \textbf{+4.9 [+3.5, +6.2]} & \textbf{+3.9 [+2.9, +5.1]} & \textbf{+3.1 [+1.9, +4.4]} & \textbf{+2.1 [+1.3, +3.0]} & \textbf{+1.9 [+0.9, +3.1]} \\
 & base & mmBERT-transplant & \textbf{+4.9 [+3.5, +6.2]} & \textbf{+4.4 [+3.4, +5.5]} & \textbf{+4.0 [+2.7, +5.2]} & \textbf{+2.6 [+1.7, +3.4]} & \textbf{+2.4 [+1.3, +3.5]} \\
 & base & mmBERT-keep & -0.1 [-1.4, +1.3] & -0.6 [-1.7, +0.7] & -1.2 [-2.6, +0.2] & \textbf{-2.3 [-3.4, -1.2]} & \textbf{-2.5 [-3.8, -1.2]} \\
 & large & Ettin & \textbf{+4.9 [+3.6, +6.0]} & \textbf{+2.9 [+1.9, +4.0]} & \textbf{+2.3 [+1.1, +3.5]} & \textbf{+1.4 [+0.6, +2.2]} & \textbf{+1.3 [+0.5, +2.1]} \\
 & 1B & Ettin & \textbf{+4.5 [+3.3, +5.6]} & +0.9 [-0.9, +2.8] & \textbf{+2.0 [+1.0, +3.1]} & \textbf{+1.3 [+0.3, +2.2]} & -- \\
\midrule
NER & xs & Ettin & \textbf{+1.9 [+1.2, +2.6]} & \textbf{+1.2 [+0.6, +1.9]} & +0.8 [-0.02, +1.6] & +0.6 [-0.04, +1.1] & 0.0 [-0.6, +0.6] \\
 & xs & Ettin-body & +0.7 [-0.04, +1.4] & +0.2 [-0.4, +0.9] & -0.1 [-0.9, +0.6] & +0.2 [-0.4, +0.8] & -0.4 [-1.0, +0.2] \\
 & s & Ettin & \textbf{+1.8 [+1.1, +2.5]} & \textbf{+1.8 [+1.2, +2.3]} & \textbf{+0.8 [+0.1, +1.4]} & \textbf{+0.7 [+0.2, +1.2]} & 0.0 [-0.5, +0.6] \\
 & s & Ettin-body & \textbf{+1.1 [+0.4, +1.8]} & \textbf{+0.9 [+0.2, +1.7]} & +0.1 [-0.6, +0.7] & +0.2 [-0.3, +0.7] & -0.5 [-1.0, +0.1] \\
 & m & Ettin & \textbf{+2.7 [+1.9, +3.4]} & \textbf{+1.6 [+0.9, +2.4]} & \textbf{+0.9 [+0.3, +1.4]} & \textbf{+0.6 [+0.1, +1.1]} & 0.0 [-0.5, +0.5] \\
 & m & Ettin-body & \textbf{+1.9 [+1.1, +2.7]} & \textbf{+0.8 [+0.2, +1.6]} & +0.6 [-0.1, +1.2] & \textbf{+0.6 [+0.1, +1.1]} & 0.0 [-0.6, +0.5] \\
 & base & Ettin & \textbf{+2.1 [+1.4, +2.8]} & \textbf{+1.4 [+0.6, +2.1]} & \textbf{+0.8 [+0.2, +1.3]} & \textbf{+0.7 [+0.2, +1.1]} & +0.4 [-0.1, +0.8] \\
 & base & mmBERT-transplant & \textbf{+2.8 [+2.1, +3.5]} & \textbf{+1.7 [+1.0, +2.3]} & \textbf{+0.8 [+0.3, +1.4]} & \textbf{+0.9 [+0.4, +1.4]} & \textbf{+0.6 [+0.1, +1.2]} \\
 & base & mmBERT-keep & \textbf{+2.9 [+2.2, +3.6]} & \textbf{+1.9 [+1.2, +2.5]} & \textbf{+1.2 [+0.5, +1.7]} & \textbf{+1.1 [+0.6, +1.6]} & \textbf{+0.8 [+0.3, +1.3]} \\
 & large & Ettin & \textbf{+1.9 [+1.3, +2.5]} & \textbf{+1.0 [+0.4, +1.4]} & \textbf{+0.7 [+0.3, +1.3]} & +0.4 [-0.1, +0.9] & 0.0 [-0.4, +0.4] \\
 & 1B & Ettin & \textbf{+1.8 [+1.2, +2.3]} & \textbf{+0.8 [+0.3, +1.3]} & +0.4 [-0.1, +0.8] & +0.3 [-0.2, +0.7] & -- \\
\midrule
ScaLA & xs & Ettin & \textbf{+8.4 [+2.6, +14.2]} & \textbf{+6.0 [+0.8, +11.5]} & +2.7 [-3.2, +8.5] & +2.7 [-1.8, +6.9] & -3.4 [-7.9, +0.9] \\
 & xs & Ettin-body & +0.7 [-7.1, +8.5] & +1.0 [-4.9, +7.3] & -1.1 [-7.3, +4.4] & +1.2 [-4.3, +6.4] & -5.0 [-10.5, +0.2] \\
 & s & Ettin & \textbf{+9.6 [+4.4, +15.2]} & +4.8 [-0.3, +9.5] & +4.7 [-0.7, +10.1] & +0.9 [-3.9, +5.1] & -1.2 [-5.8, +3.1] \\
 & s & Ettin-body & \textbf{+7.4 [+1.6, +12.7]} & +1.3 [-3.7, +6.2] & +0.6 [-4.6, +5.9] & +1.7 [-2.9, +6.9] & -0.4 [-5.1, +4.2] \\
 & m & Ettin & \textbf{+8.3 [+3.5, +13.3]} & \textbf{+6.6 [+2.0, +10.9]} & +4.1 [-0.3, +8.4] & +2.4 [-2.3, +6.7] & -0.9 [-5.5, +3.5] \\
 & m & Ettin-body & \textbf{+9.7 [+4.0, +14.8]} & \textbf{+8.1 [+3.2, +13.0]} & +3.7 [-1.0, +8.4] & +1.8 [-2.9, +6.5] & -1.5 [-6.3, +2.9] \\
 & base & Ettin & \textbf{+8.2 [+3.3, +13.1]} & \textbf{+6.6 [+1.9, +11.0]} & +1.7 [-2.7, +5.5] & \textbf{+4.7 [+1.4, +8.4]} & +2.3 [-1.3, +5.9] \\
 & base & mmBERT-transplant & \textbf{+13.3 [+7.8, +18.3]} & \textbf{+9.6 [+5.1, +14.4]} & \textbf{+5.1 [+1.1, +8.8]} & \textbf{+5.5 [+2.1, +8.3]} & +3.0 [-0.2, +6.5] \\
 & base & mmBERT-keep & \textbf{+15.7 [+10.0, +21.1]} & \textbf{+12.1 [+7.2, +17.0]} & \textbf{+4.4 [+0.02, +8.7]} & +2.5 [-1.5, +6.4] & +0.1 [-4.1, +4.3] \\
 & large & Ettin & \textbf{+9.0 [+4.4, +13.6]} & \textbf{+5.6 [+1.7, +9.9]} & +2.0 [-1.5, +5.7] & +0.2 [-2.8, +3.2] & \textbf{+3.3 [+0.1, +7.1]} \\
 & 1B & Ettin & \textbf{+11.8 [+7.5, +16.0]} & \textbf{+8.8 [+5.3, +12.6]} & \textbf{+6.9 [+4.0, +9.9]} & +2.3 [-0.5, +5.0] & -- \\
\midrule
IceEC & xs & Ettin & +2.2 [-0.01, +4.4] & +0.2 [-2.1, +2.3] & -0.9 [-3.2, +1.3] & -0.4 [-2.0, +1.2] & -0.9 [-2.7, +0.8] \\
 & xs & Ettin-body & +2.1 [-0.2, +4.2] & +1.5 [-0.6, +3.5] & +0.7 [-1.6, +3.0] & +0.9 [-1.2, +3.2] & +0.4 [-1.8, +2.8] \\
 & s & Ettin & \textbf{+2.6 [+0.8, +4.3]} & +1.5 [-0.3, +3.4] & +0.5 [-2.0, +2.9] & -1.4 [-3.1, +0.3] & \textbf{-2.0 [-3.9, -0.04]} \\
 & s & Ettin-body & +0.9 [-1.5, +3.0] & -0.7 [-2.6, +1.3] & +0.3 [-1.9, +2.2] & -1.5 [-3.0, +0.1] & \textbf{-2.1 [-3.9, -0.01]} \\
 & m & Ettin & \textbf{+3.7 [+1.8, +5.7]} & +2.0 [-0.1, +3.9] & +2.2 [-0.2, +4.7] & +1.7 [-0.2, +3.7] & -0.2 [-2.0, +1.6] \\
 & m & Ettin-body & \textbf{+4.0 [+1.8, +6.1]} & +1.7 [-0.6, +3.9] & \textbf{+3.1 [+0.9, +5.4]} & +1.7 [-0.4, +3.7] & -0.2 [-2.1, +1.6] \\
 & base & Ettin & \textbf{+4.1 [+1.8, +6.3]} & \textbf{+3.1 [+0.5, +5.7]} & +0.2 [-2.0, +2.4] & +1.2 [-0.8, +3.3] & +0.3 [-1.8, +2.6] \\
 & base & mmBERT-transplant & \textbf{+5.5 [+1.9, +8.7]} & \textbf{+4.4 [+1.3, +7.5]} & +0.7 [-1.5, +3.1] & \textbf{+2.6 [+0.7, +4.6]} & +1.8 [-0.4, +4.1] \\
 & base & mmBERT-keep & \textbf{+5.3 [+2.9, +7.6]} & \textbf{+6.1 [+3.0, +9.0]} & \textbf{+3.3 [+1.0, +5.5]} & \textbf{+2.3 [+0.3, +4.1]} & +1.4 [-0.7, +3.6] \\
 & large & Ettin & \textbf{+2.8 [+0.5, +5.1]} & \textbf{+3.9 [+1.9, +6.1]} & +1.9 [-0.3, +4.2] & +0.9 [-0.8, +2.7] & -0.1 [-1.9, +1.7] \\
 & 1B & Ettin & \textbf{+6.8 [+4.7, +9.3]} & \textbf{+3.6 [+1.8, +5.5]} & \textbf{+2.8 [+0.8, +4.8]} & +1.5 [-0.4, +3.1] & -- \\
\midrule
PoS & xs & Ettin & -- & -- & -- & 0.0 [-0.4, +0.4] & +0.1 [-0.3, +0.4] \\
 & xs & Ettin-body & -- & -- & -- & -0.2 [-0.4, +0.1] & -0.1 [-0.3, +0.1] \\
 & s & Ettin & -- & -- & -- & \textbf{+0.3 [+0.1, +0.5]} & +0.2 [-0.1, +0.4] \\
 & s & Ettin-body & -- & -- & -- & \textbf{+0.3 [+0.05, +0.5]} & +0.2 [-0.04, +0.4] \\
 & m & Ettin & -- & -- & -- & \textbf{+0.2 [+0.002, +0.5]} & +0.2 [-0.1, +0.4] \\
 & m & Ettin-body & -- & -- & -- & \textbf{+0.3 [+0.2, +0.5]} & \textbf{+0.3 [+0.1, +0.5]} \\
 & base & Ettin & -- & -- & -- & \textbf{+0.3 [+0.1, +0.4]} & \textbf{+0.2 [+0.04, +0.4]} \\
 & base & mmBERT-transplant & -- & -- & -- & +0.1 [-0.1, +0.2] & 0.0 [-0.1, +0.2] \\
 & base & mmBERT-keep & -- & -- & -- & +0.2 [-0.01, +0.4] & +0.1 [-0.1, +0.3] \\
 & large & Ettin & -- & -- & -- & \textbf{+0.2 [+0.03, +0.4]} & +0.2 [-0.03, +0.3] \\
 & 1B & Ettin & -- & -- & -- & \textbf{+0.2 [+0.04, +0.3]} & -- \\
\bottomrule
\end{tabular}}
\caption{Warm start minus scratch ($\times$100) at the same size and token budget, and warm start at 80 tokens per parameter minus scratch at 160, with paired 95\% confidence intervals (hierarchical bootstrap over pretraining seeds, fine-tuning seeds and test items); bold where the interval excludes zero (bounds that round to 0.0 are given to two decimals). Two pretraining seeds per arm at xs, s and m (one for Ettin-body), one at base, large and 1B. PoS is measured at 80 tokens per parameter only; 1B has no scratch model at 160. The 170 same-budget differences counted in Section~\ref{sec:res-warm} are the same-budget cells of Ettin, Ettin-body and mmBERT-transplant on the five tasks.}
\label{tab:warm-diff}
\end{table*}

%% file: tables/appendix_warm160.tex
\begin{table*}[t]
\centering
\small
\setlength{\tabcolsep}{2.2pt}
\begin{tabular}{lllcccc}
\toprule
Task & Size & Arm & 10 & 20 & 40 & 80 \\
\midrule
QA & xs & Ettin & \textbf{+1.6 [+0.5, +2.6]} & \textbf{+2.0 [+0.9, +3.2]} & \textbf{+2.3 [+1.2, +3.4]} & \textbf{+2.2 [+0.8, +3.5]} \\
 & xs & Ettin-body & \textbf{-2.7 [-5.3, -0.7]} & -0.3 [-1.7, +1.1] & \textbf{+1.8 [+0.3, +3.2]} & \textbf{+1.4 [+0.2, +2.8]} \\
 & s & Ettin & +0.4 [-0.7, +1.4] & 0.0 [-1.3, +1.2] & +0.5 [-0.6, +1.5] & \textbf{+1.4 [+0.4, +2.3]} \\
 & s & Ettin-body & -0.7 [-1.9, +0.5] & -0.5 [-1.8, +0.7] & -1.2 [-3.2, +0.5] & +0.6 [-0.5, +1.9] \\
 & m & Ettin & +0.9 [-0.1, +2.0] & +1.3 [-0.1, +2.6] & \textbf{+1.0 [+0.1, +2.1]} & \textbf{+1.3 [+0.1, +2.6]} \\
 & m & Ettin-body & +0.5 [-0.5, +1.6] & +0.7 [-0.4, +1.8] & \textbf{+0.9 [+0.02, +1.8]} & \textbf{+1.3 [+0.3, +2.2]} \\
 & base & Ettin & \textbf{+1.9 [+0.8, +3.1]} & \textbf{+1.6 [+0.4, +2.9]} & \textbf{+2.0 [+0.8, +3.2]} & \textbf{+1.9 [+0.9, +3.1]} \\
 & base & mmBERT-transplant & \textbf{+1.9 [+0.8, +3.1]} & \textbf{+2.1 [+1.0, +3.3]} & \textbf{+2.8 [+1.7, +4.0]} & \textbf{+2.4 [+1.3, +3.5]} \\
 & base & mmBERT-keep & \textbf{-3.0 [-4.2, -1.7]} & \textbf{-2.9 [-4.1, -1.6]} & \textbf{-2.4 [-3.6, -1.0]} & \textbf{-2.5 [-3.8, -1.2]} \\
 & large & Ettin & +0.7 [-0.4, +1.7] & +0.5 [-0.4, +1.4] & \textbf{+1.4 [+0.4, +2.5]} & \textbf{+1.3 [+0.5, +2.1]} \\
\midrule
NER & xs & Ettin & \textbf{-0.8 [-1.4, -0.2]} & -0.4 [-1.0, +0.1] & -0.2 [-0.8, +0.4] & 0.0 [-0.6, +0.6] \\
 & xs & Ettin-body & \textbf{-1.9 [-2.8, -1.2]} & \textbf{-1.4 [-2.2, -0.8]} & \textbf{-1.1 [-1.8, -0.5]} & -0.4 [-1.0, +0.2] \\
 & s & Ettin & \textbf{-0.9 [-1.5, -0.3]} & -0.3 [-0.9, +0.3] & -0.3 [-1.0, +0.4] & 0.0 [-0.5, +0.6] \\
 & s & Ettin-body & \textbf{-1.6 [-2.2, -0.9]} & \textbf{-1.1 [-1.9, -0.4]} & \textbf{-1.0 [-1.7, -0.3]} & -0.5 [-1.0, +0.1] \\
 & m & Ettin & \textbf{-0.7 [-1.2, -0.2]} & -0.3 [-0.9, +0.2] & -0.4 [-0.9, +0.1] & 0.0 [-0.5, +0.5] \\
 & m & Ettin-body & \textbf{-1.4 [-2.0, -0.7]} & \textbf{-1.1 [-1.7, -0.6]} & \textbf{-0.7 [-1.3, -0.1]} & 0.0 [-0.6, +0.5] \\
 & base & Ettin & \textbf{-0.7 [-1.2, -0.2]} & -0.1 [-0.7, +0.5] & +0.2 [-0.2, +0.7] & +0.4 [-0.1, +0.8] \\
 & base & mmBERT-transplant & 0.0 [-0.5, +0.5] & +0.3 [-0.3, +0.8] & +0.3 [-0.2, +0.8] & \textbf{+0.6 [+0.1, +1.2]} \\
 & base & mmBERT-keep & +0.1 [-0.5, +0.7] & \textbf{+0.5 [+0.001, +1.0]} & \textbf{+0.6 [+0.1, +1.1]} & \textbf{+0.8 [+0.3, +1.3]} \\
 & large & Ettin & \textbf{-0.5 [-1.0, -0.1]} & \textbf{-0.5 [-0.9, -0.02]} & -0.2 [-0.7, +0.2] & 0.0 [-0.4, +0.4] \\
\midrule
ScaLA & xs & Ettin & \textbf{-12.1 [-17.3, -6.7]} & \textbf{-8.9 [-13.5, -4.5]} & \textbf{-6.2 [-11.8, -0.9]} & -3.4 [-7.9, +0.9] \\
 & xs & Ettin-body & \textbf{-19.9 [-27.3, -13.3]} & \textbf{-13.9 [-19.8, -7.7]} & \textbf{-10.1 [-15.4, -5.0]} & -5.0 [-10.5, +0.2] \\
 & s & Ettin & \textbf{-8.5 [-13.0, -3.4]} & \textbf{-6.2 [-10.6, -1.6]} & -2.0 [-6.1, +2.4] & -1.2 [-5.8, +3.1] \\
 & s & Ettin-body & \textbf{-10.8 [-15.7, -5.9]} & \textbf{-9.7 [-14.2, -5.2]} & \textbf{-6.1 [-10.7, -1.9]} & -0.4 [-5.1, +4.2] \\
 & m & Ettin & \textbf{-7.9 [-12.3, -3.6]} & \textbf{-5.7 [-9.7, -1.8]} & -1.3 [-5.1, +2.6] & -0.9 [-5.5, +3.5] \\
 & m & Ettin-body & \textbf{-6.5 [-10.9, -1.8]} & \textbf{-4.2 [-8.8, -0.03]} & -1.7 [-6.0, +2.5] & -1.5 [-6.3, +2.9] \\
 & base & Ettin & \textbf{-9.1 [-13.9, -4.8]} & -4.3 [-8.7, +0.2] & -1.2 [-5.4, +2.9] & +2.3 [-1.3, +5.9] \\
 & base & mmBERT-transplant & \textbf{-4.1 [-8.6, -0.1]} & -1.3 [-5.4, +2.9] & +2.2 [-1.8, +5.8] & +3.0 [-0.2, +6.5] \\
 & base & mmBERT-keep & -1.6 [-6.0, +2.5] & +1.2 [-3.1, +5.5] & +1.5 [-2.8, +5.8] & +0.1 [-4.1, +4.3] \\
 & large & Ettin & -2.1 [-6.4, +2.3] & 0.0 [-3.5, +4.1] & +2.0 [-1.8, +5.8] & \textbf{+3.3 [+0.1, +7.1]} \\
\midrule
IceEC & xs & Ettin & \textbf{-2.0 [-4.1, -0.1]} & -2.2 [-4.6, +0.04] & \textbf{-1.9 [-3.9, -0.1]} & -0.9 [-2.7, +0.8] \\
 & xs & Ettin-body & \textbf{-2.2 [-4.3, -0.04]} & -0.8 [-2.9, +1.3] & -0.3 [-2.3, +1.7] & +0.4 [-1.8, +2.8] \\
 & s & Ettin & \textbf{-2.9 [-4.8, -0.8]} & -1.4 [-3.5, +0.8] & -2.1 [-4.2, +0.2] & \textbf{-2.0 [-3.9, -0.04]} \\
 & s & Ettin-body & \textbf{-4.6 [-7.0, -2.2]} & \textbf{-3.6 [-5.8, -1.5]} & \textbf{-2.2 [-4.3, -0.1]} & \textbf{-2.1 [-3.9, -0.01]} \\
 & m & Ettin & \textbf{-2.8 [-4.5, -1.0]} & -1.6 [-3.6, +0.2] & -0.9 [-3.0, +1.2] & -0.2 [-2.0, +1.6] \\
 & m & Ettin-body & \textbf{-2.5 [-4.5, -0.4]} & -1.9 [-3.9, +0.1] & -0.1 [-1.9, +1.8] & -0.2 [-2.1, +1.6] \\
 & base & Ettin & -1.9 [-4.3, +0.3] & -1.0 [-3.1, +1.1] & -0.6 [-2.9, +1.5] & +0.3 [-1.8, +2.6] \\
 & base & mmBERT-transplant & -0.5 [-4.0, +2.6] & +0.3 [-2.4, +3.0] & -0.1 [-2.5, +2.4] & +1.8 [-0.4, +4.1] \\
 & base & mmBERT-keep & -0.7 [-3.0, +1.7] & +2.0 [-0.5, +4.5] & \textbf{+2.4 [+0.1, +4.8]} & +1.4 [-0.7, +3.6] \\
 & large & Ettin & \textbf{-3.9 [-5.8, -1.8]} & -1.5 [-3.5, +0.2] & -1.3 [-3.5, +1.0] & -0.1 [-1.9, +1.7] \\
\bottomrule
\end{tabular}
\caption{Warm start at each budget (tokens per parameter) minus scratch at 160 tokens per parameter ($\times$100), with paired 95\% confidence intervals; bold where the interval excludes zero (bounds that round to 0.0 are given to two decimals). These are the comparisons behind the data savings in Section~\ref{sec:res-warm}.}
\label{tab:warm-diff160}
\end{table*}

%% file: tables/appendix_checks.tex
\begin{table}[t]
\centering
\small
\setlength{\tabcolsep}{3pt}
\resizebox{\columnwidth}{!}{%
\begin{tabular}{llcc}
\toprule
Task & & reused rate & re-tuned rate \\
\midrule
QA & Scratch & 1e-4: 51.6 & 1e-4: 51.6 \\
 & Ettin & 1e-4: 56.5 & 4e-4: 55.8 \\
 & Ettin $-$ Scratch & \textbf{+4.9 [+3.5, +6.3]} & \textbf{+4.2 [+2.8, +5.5]} \\
ScaLA & Scratch & 6e-5: 52.7 & 3e-5: 56.3 \\
 & Ettin & 1e-4: 60.9 & 3e-5: 62.9 \\
 & Ettin $-$ Scratch & \textbf{+8.2 [+3.3, +13.1]} & \textbf{+6.5 [+2.1, +11.0]} \\
\bottomrule
\end{tabular}}
\caption{Re-tuning the fine-tuning learning rate at 10 tokens per parameter (base size): the rate and test score ($\times$100, mean of three seeds) with the rate reused from the 80-token model and with the rate chosen on development data from the 80-token model's grid, and the paired difference with its 95\% interval (bold: excludes zero).}
\label{tab:check-lr}
\end{table}

\begin{table}[t]
\centering
\small
\setlength{\tabcolsep}{3pt}
\resizebox{\columnwidth}{!}{%
\begin{tabular}{lccccc}
\toprule
Model & 1k & 2k & 4k & 8k & 1k $-$ 8k \\
\midrule
xs, Scratch & 2.533 & 2.464 & 2.462 & 2.470 & +0.063 [+0.041, +0.089] \\
s, Scratch & 2.090 & 2.025 & 2.009 & 2.016 & +0.075 [+0.054, +0.099] \\
m, Scratch & 1.673 & 1.620 & 1.602 & 1.604 & +0.069 [+0.051, +0.089] \\
base, Scratch & 1.399 & 1.360 & 1.349 & 1.351 & +0.048 [+0.034, +0.066] \\
large, Scratch & 1.117 & 1.076 & 1.064 & 1.066 & +0.051 [+0.038, +0.067] \\
1B, Scratch & 0.884 & 0.855 & 0.846 & 0.847 & +0.037 [+0.028, +0.048] \\
xs, Ettin & 2.358 & 2.284 & 2.263 & 2.269 & +0.088 [+0.064, +0.115] \\
s, Ettin & 1.929 & 1.864 & 1.844 & 1.845 & +0.084 [+0.063, +0.108] \\
m, Ettin & 1.523 & 1.468 & 1.446 & 1.441 & +0.082 [+0.062, +0.103] \\
base, Ettin & 1.260 & 1.208 & 1.190 & 1.179 & +0.080 [+0.062, +0.100] \\
large, Ettin & 0.975 & 0.938 & 0.920 & 0.917 & +0.058 [+0.042, +0.075] \\
1B, Ettin & 0.773 & 0.742 & 0.730 & 0.727 & +0.046 [+0.034, +0.059] \\
\bottomrule
\end{tabular}}
\caption{Masked-token loss (nats) on the last 1,024 positions of 54 held-out documents of at least 8,192 tokens, when the input is the last 1k, 2k, 4k or all 8k tokens, for the models at 80 tokens per parameter; the last column is the gain from the full context with its 95\% interval over documents. The gain is positive in all 25 of the 25 models we ran with our vocabulary.}
\label{tab:check-longctx}
\end{table}

\begin{table*}[t]
\centering
\small
\setlength{\tabcolsep}{5pt}
\begin{tabular}{llccccc}
\toprule
Model & Task & full rate & half rate & half $-$ full & full $-$ Scratch & half $-$ Scratch \\
\midrule
Ettin & MLM loss & 1.384 & 1.557 & +0.173 & & \\
 & QA & 56.5 & 56.5 & 0.0 [-0.8, +0.9] & \textbf{+4.9 [+3.4, +6.3]} & \textbf{+4.9 [+3.6, +6.3]} \\
 & NER & 92.9 & 92.5 & -0.4 [-0.9, +0.1] & \textbf{+2.1 [+1.4, +2.8]} & \textbf{+1.7 [+1.1, +2.4]} \\
 & ScaLA & 60.9 & 54.3 & \textbf{-6.6 [-11.1, -2.1]} & \textbf{+8.2 [+3.5, +13.2]} & +1.7 [-3.4, +6.9] \\
 & IceEC & 46.3 & 43.6 & \textbf{-2.6 [-4.7, -0.7]} & \textbf{+4.1 [+2.2, +6.3]} & +1.5 [-0.9, +3.6] \\
\midrule
mmBERT-keep & MLM loss & 0.430 & 0.440 & +0.009 & & \\
 & QA & 51.5 & 50.8 & -0.7 [-1.7, +0.3] & -0.1 [-1.5, +1.3] & -0.7 [-2.1, +0.6] \\
 & NER & 93.7 & 93.3 & -0.3 [-0.8, +0.2] & \textbf{+2.9 [+2.2, +3.6]} & \textbf{+2.5 [+1.9, +3.2]} \\
 & ScaLA & 68.4 & 68.4 & 0.0 [-3.7, +3.5] & \textbf{+15.7 [+10.6, +21.0]} & \textbf{+15.8 [+10.5, +21.0]} \\
 & IceEC & 47.4 & 48.0 & +0.6 [-1.8, +3.0] & \textbf{+5.3 [+3.2, +7.4]} & \textbf{+5.9 [+3.4, +8.7]} \\
\bottomrule
\end{tabular}
\caption{Pretraining at half the scratch peak learning rate, at base size and 10 tokens per parameter: the warm-started model trained at the full rate (the study's branch) and one trained straight at half the rate, both fine-tuned at the branch's rate. MLM loss is the mean validation loss at the end (mmBERT-keep's is per Gemma token, so it compares only with itself); task scores are test $\times$100, mean of three seeds, with paired differences and 95\% intervals (bold: excludes zero). Scratch is the 10-token branch of the scratch curve.}
\label{tab:check-pretrain-lr}
\end{table*}